\documentclass{article} 
\usepackage{iclr2027_conference,times}

\usepackage{amsmath,amsfonts,bm}

\def\eqref#1{equation~\ref{#1}}

\def\1{\bm{1}}

\DeclareMathAlphabet{\mathsfit}{\encodingdefault}{\sfdefault}{m}{sl}
\SetMathAlphabet{\mathsfit}{bold}{\encodingdefault}{\sfdefault}{bx}{n}

\usepackage{hyperref}
\usepackage{url}
\usepackage[utf8]{inputenc}
\usepackage[T1]{fontenc}
\usepackage{microtype}
\usepackage{booktabs}
\usepackage{colortbl}
\usepackage{amsmath}
\usepackage{amssymb}
\usepackage{amsfonts}
\usepackage{nicefrac}
\usepackage{graphicx}
\usepackage{xcolor}
\usepackage{multirow}
\usepackage{subcaption}
\usepackage{enumitem}
\usepackage{wrapfig}
\usepackage{comment}
\usepackage{color}
\usepackage{booktabs}
\usepackage{booktabs} 
\usepackage{amssymb}  
\usepackage{xcolor}   
\usepackage{makecell} 
\usepackage{algpseudocode}
\definecolor{darkblue}{rgb}{0, 0, 0.5}
\hypersetup{colorlinks=true, citecolor=darkblue, linkcolor=darkblue, urlcolor=darkblue}
\usepackage[most]{tcolorbox}
\usepackage{enumitem}
\usepackage{pifont,xcolor,booktabs,array}
\usepackage{wasysym}
\usepackage{fontawesome5}
\usepackage{academicons}
\usepackage{svg}
\usepackage{algorithm}
\usepackage{algpseudocode}
\usepackage{fontawesome5} 

\usepackage{xspace} 
\newcommand{\ie}{{\sl i.e.}}
\newcommand{\eg}{{\sl e.g.}}

\newcommand{\github}{\raisebox{-1.5pt}{\includegraphics[height=1.05em]{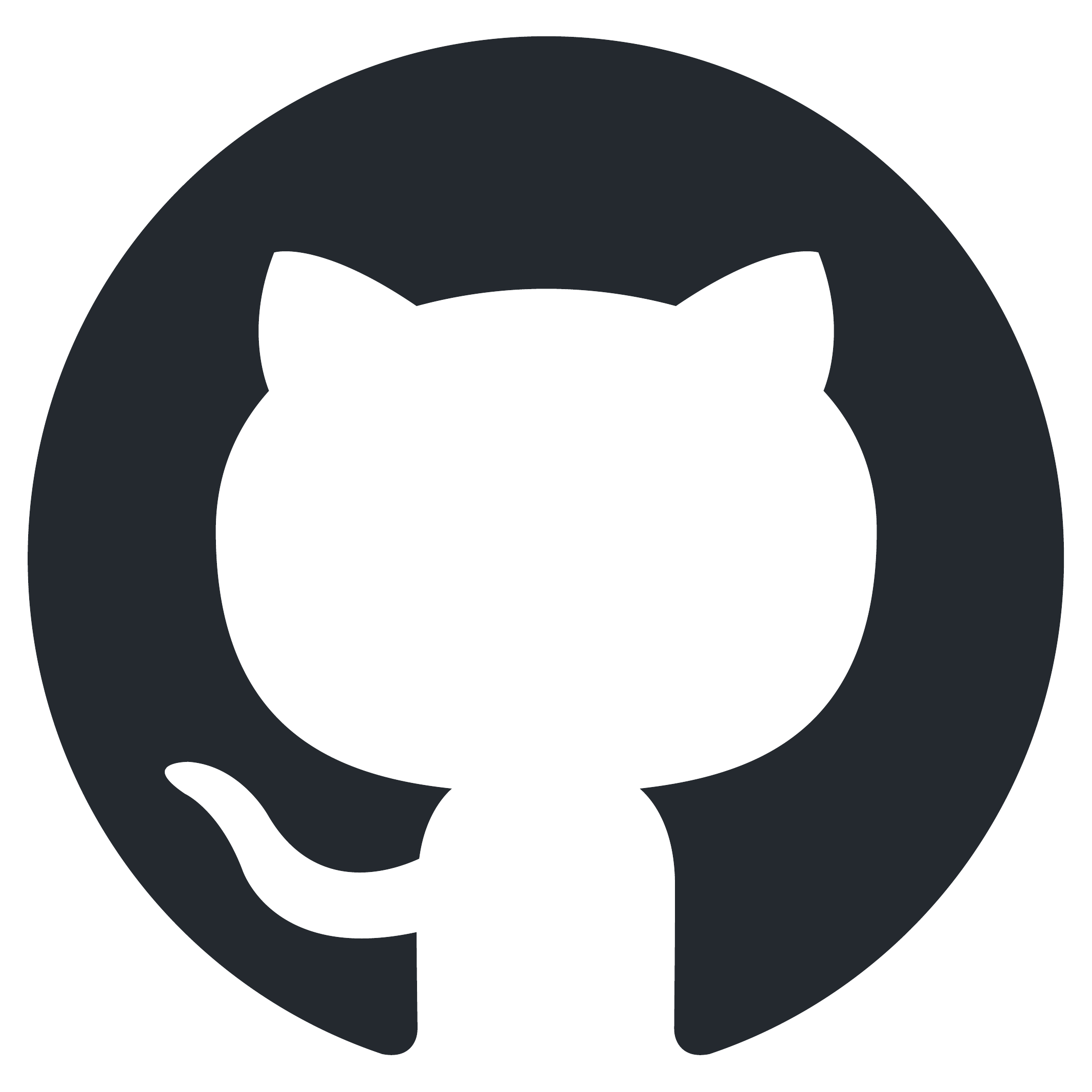}}\xspace}
\newcommand{\huggingface}{\raisebox{-1.5pt}{\includegraphics[height=1.05em]{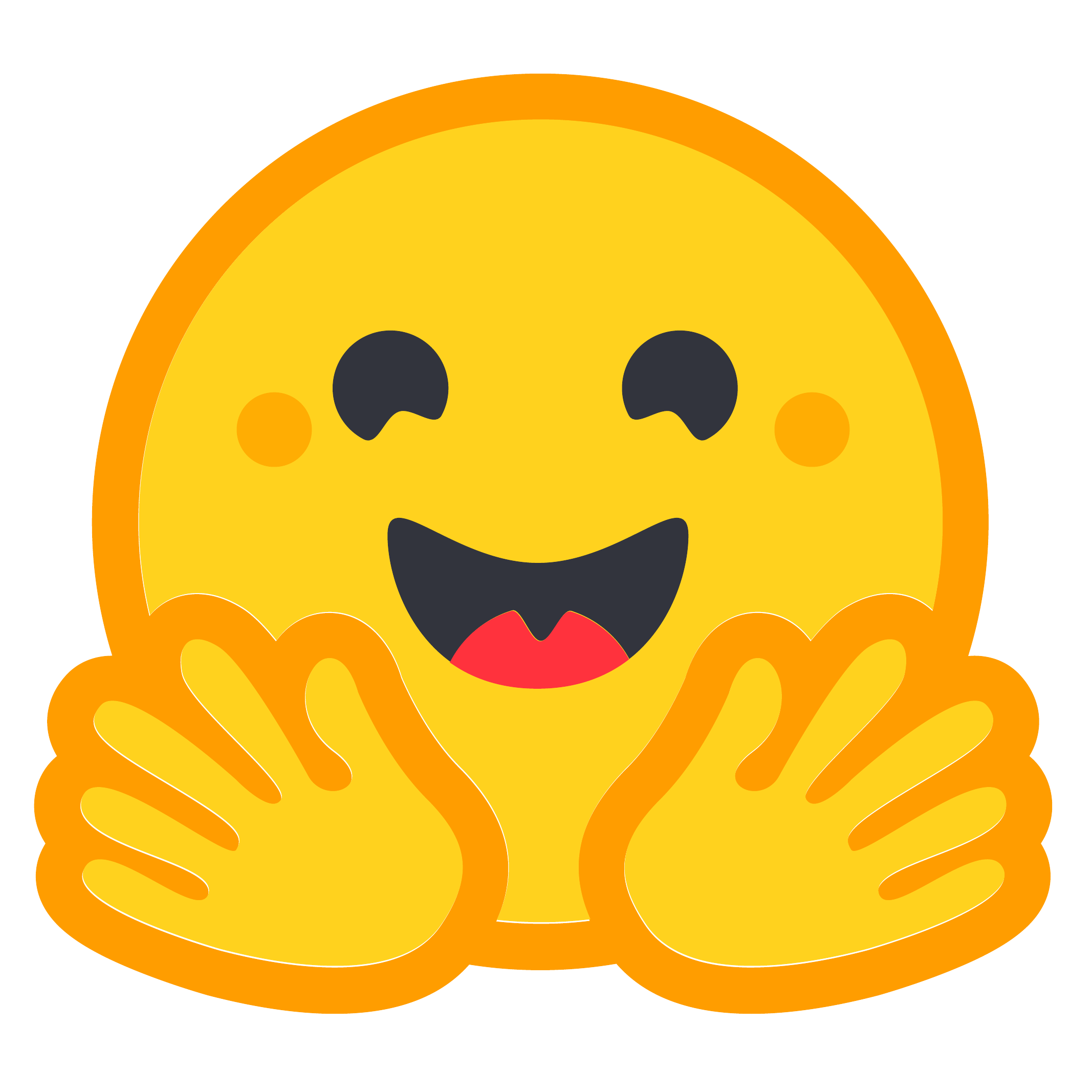}}\xspace}

\newcommand{\good}{\textcolor{green!60!black}{\smiley}}
\newcommand{\bad}{\textcolor{red!70!black}{\frownie}}

\definecolor{TakeawayFrame}{HTML}{003366} 
\definecolor{TakeawayBodyBG}{HTML}{F0F4F8} 
\definecolor{TakeawayTitleText}{HTML}{FFFFFF}

\newtcolorbox{takeawaybox}[1]{
    enhanced,
    colback=TakeawayBodyBG,
    colframe=TakeawayFrame,
    coltitle=TakeawayTitleText,
    fonttitle=\bfseries,
    title=#1,
    boxrule=1.2pt,
    arc=4pt,
    left=10pt, right=10pt, top=8pt, bottom=8pt,
    drop fuzzy shadow=black!15,
    before upper={\setlist[enumerate]{label=\arabic*., leftmargin=*, itemsep=4pt, parsep=0pt, topsep=4pt}}
}

\title{Negative Self-Distillation: Learning to\\Reason by Avoiding Flaws}

\author{
Rongcan Pei$^1$, Zhepei Wei$^1$, Shuyao Xu$^2$, Xinyu Zhu$^1$, Wei-Lin Chen$^1$, and Yu Meng$^1$ \\
$^1$Department of Computer Science, University of Virginia \quad $^2$Stanford University\\
\texttt{\{peirongcan,zhepei.wei,xinyuzhu,wlchen,yumeng5\}@virginia.edu} \\ \texttt{{shuyao}@stanford.edu}\\[4pt]
\href{https://github.com/Prongcan/NSD}{\github\ GitHub}
\hspace{12pt}
\href{https://huggingface.co/collections/PassionPrc/nsd-negative-self-distillation}{
\huggingface
\ Hugging Face
}
}

\iclrfinalcopy 
\begin{document}
\maketitle

\vspace{-5mm}
\begin{abstract}

On-Policy Self-Distillation (OPSD) has emerged as a popular paradigm for large language model (LLM) self-improvement, allowing models to act as their own teachers by leveraging privileged information such as ground-truth solutions. 
However, recent findings indicate that OPSD can severely degrade the performance of LLMs on complex reasoning tasks: By forcing the student to imitate an artificially confident reasoning trace conditioned on privileged information, OPSD inadvertently suppresses expressions of uncertainty and penalizes the exploratory, self-corrective behaviors required to solve challenging problems. 
To address this, we introduce Negative Self-Distillation (NSD), a new framework that optimizes LLMs by diverging from flawed reasoning rather than imitating privileged solutions.
Instead of relying on ground-truth answers or external supervision, NSD uses the model itself to generate a question-specific negative condition (\eg, acting as a ``careless reasoner'') and pushes the student's distribution away from this self-generated negative teacher.
Naively applying unlearning objectives to achieve this divergence is problematic, as flawed reasoning tokens are confounded with basic linguistic tokens; indiscriminately penalizing both risks catastrophically degrading the model's foundational language capabilities.
We resolve this by designing a dynamic gating mechanism that automatically identifies and isolates reasoning-critical tokens, ensuring gradient updates target only behavioral flaws while preserving the model's linguistic priors.
Empirically, NSD consistently outperforms OPSD and other label-free, self-bootstrapping reinforcement learning (RL) baselines. 
Across seven mathematical reasoning benchmarks (AIME 24/25/26, HMMT, AMC, OlympiadBench, and MATH), NSD achieves average gains of 2.3\%, 7.5\%, and 6.0\% for 1.7B, 4B, and 8B models, respectively.
Further analyses show that NSD achieves higher training efficiency while preserving the self-correction behaviors crucial for complex reasoning.

\end{abstract}
\vspace{-5mm}

\begin{figure}[htbp]
    \centering
    \includegraphics[width=1\textwidth]{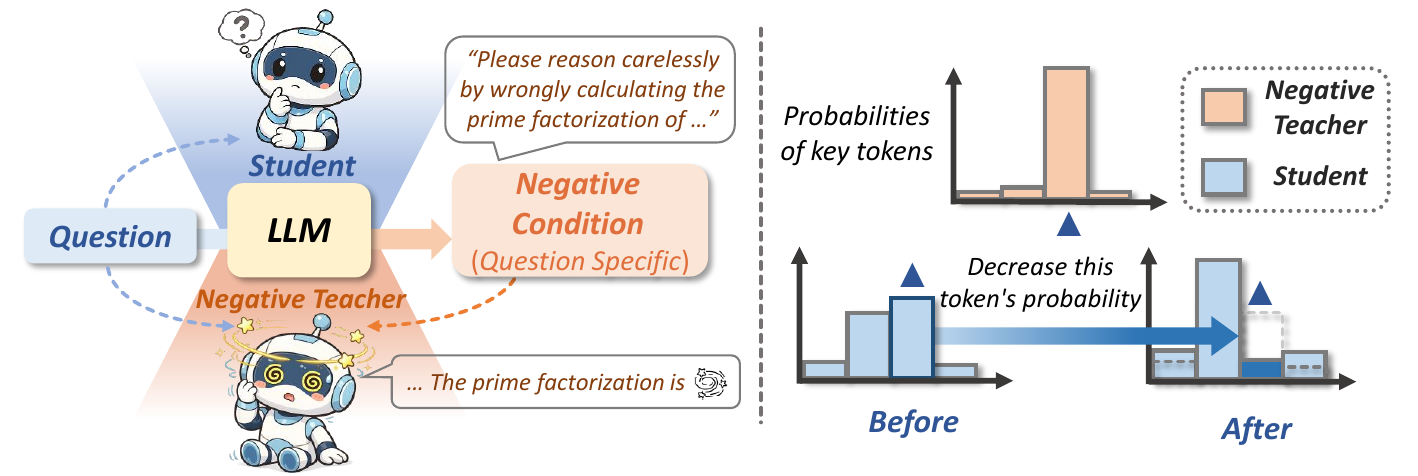} 
    \caption{Overview of the NSD framework. (Left) We construct a negative teacher from the same base model via self-generated negative conditioning. (Right) The student model is optimized to diverge its distribution from that of the negative teacher.}
    \label{fig:intro}
\end{figure}
\section{Introduction}
\label{sec:intro}


Reinforcement Learning with Verifiable Rewards (RLVR)~\citep{grpo,dapo,tulu} has emerged as an effective paradigm for enhancing the reasoning capabilities of large language models (LLMs). 
However, RLVR is often bottlenecked by computational inefficiency and training signal sparsity. 
These challenges arise because (1) sampling multiple rollouts per query is expensive, and rollouts within a group frequently receive identical rewards on exceptionally easy or difficult problems, leading to advantage collapse and vanishing gradients~\citep{grposparse,arrol,EDGE-GRPO}; 
and (2) outcome-based rewards are applied uniformly across the entire generated sequence, which obscures fine-grained, token-level credit assignment. 
To mitigate these limitations, On-Policy Distillation (OPD)~\citep{opd,lu2025onpolicydistillation,opd2026survey} utilizes a stronger, external teacher model to provide dense token-level supervision over the student model's self-sampled reasoning trajectories. 
While this approach successfully yields richer feedback, it introduces a practical constraint:
obtaining a strictly superior external teacher that is both sufficiently capable of providing accurate dense supervision and compatible with the student's tokenizer is often impractical.


To circumvent the reliance on external teacher models, On-Policy Self-Distillation (OPSD)~\citep{opsd,sdft,sdpo} has been proposed as a scalable alternative. 
In OPSD, the model acts as its own teacher by utilizing privileged information (\eg, ground-truth answers) to generate dense supervision signals for the student's self-sampled trajectories. 
However, because the OPSD teacher inherently knows the ground-truth solution, it tends to produce artificially confident and highly linear reasoning trajectories~\citep{opsdfail,PI-biased}.
Consequently, forcing the student to minimize the divergence from this teacher distribution inadvertently suppresses high-entropy exploration, expressions of uncertainty, and self-corrective behaviors, which are essential for complex problem-solving.


Motivated by the observation that imitating a synthetically confident oracle can degrade natural reasoning processes, we explore an alternative training paradigm: optimizing the model to explicitly avoid flawed reasoning patterns. 
We introduce \textbf{Negative Self-Distillation (NSD)}, a fully self-bootstrapped framework that operates without external privileged data. 
Instead of utilizing a teacher conditioned on the correct answer, the model is prompted to generate a question-specific negative condition (\eg, acting as a ``careless reasoner'') to instantiate a negative teacher. 
The student is then optimized to move its token distribution away from the negative teacher, encouraging it to avoid premature conclusions and other flawed reasoning patterns. 
Importantly, the negative signal is generated from the model itself and does not require ground-truth solutions or external annotations. 

A central challenge, however, is that not every token assigned high likelihood by the negatively conditioned teacher corresponds to a reasoning error.
A naive divergence or unlikelihood objective~\citep{unlikelihood} can also penalize ordinary linguistic tokens, degrading the model's pretrained linguistic priors. 
NSD therefore introduces a dynamic token-level gating mechanism that compares the negative teacher with a benign reference model and activates the negative objective only when the negative condition increases the likelihood of the sampled token. 
We further stabilize these updates with a bounded unlikelihood formulation and a KL-based regularization term, preventing excessive updates on high-confidence structural tokens while retaining targeted supervision on reasoning-critical tokens. 
This design yields a training signal that is both selective and computationally efficient: 
NSD requires only a single student rollout per sample, avoids full-vocabulary logit alignment, and can parallelize the reference and negative-teacher computations. 
Beyond accuracy, our analysis shows that NSD preserves and strengthens reflective self-correction behavior rather than encouraging overly confident, linear reasoning. Our main contributions are summarized as follows:

\begin{itemize}[leftmargin=1em]
    \item We propose Negative Self-Distillation (NSD), a label-free, fully self-bootstrapped framework that enhances reasoning capabilities by optimizing the model to diverge from self-generated flawed trajectories, eliminating the need for ground-truth solutions or an external teacher.
    
    \item 
    We introduce a token-level gating mechanism together with a bounded unlikelihood objective, enabling targeted divergence from flawed reasoning while preserving foundational language priors.
    
    \item 
    We demonstrate that NSD consistently outperforms existing training paradigms (\ie,  OPSD~\citep{opsd}, Intuitor~\citep{rlif} and TTRL~\citep{ttrl}) across 1.7B, 4B, and 8B model sizes on seven reasoning tasks.
    Furthermore, NSD achieves superior training efficiency, mitigates overconfidence, and preserves the model's intrinsic reflection capabilities.
\end{itemize}

\section{NSD: Negative Self-Distillation}

\begin{figure}[htbp]
    \centering
    \includegraphics[width=1\textwidth]{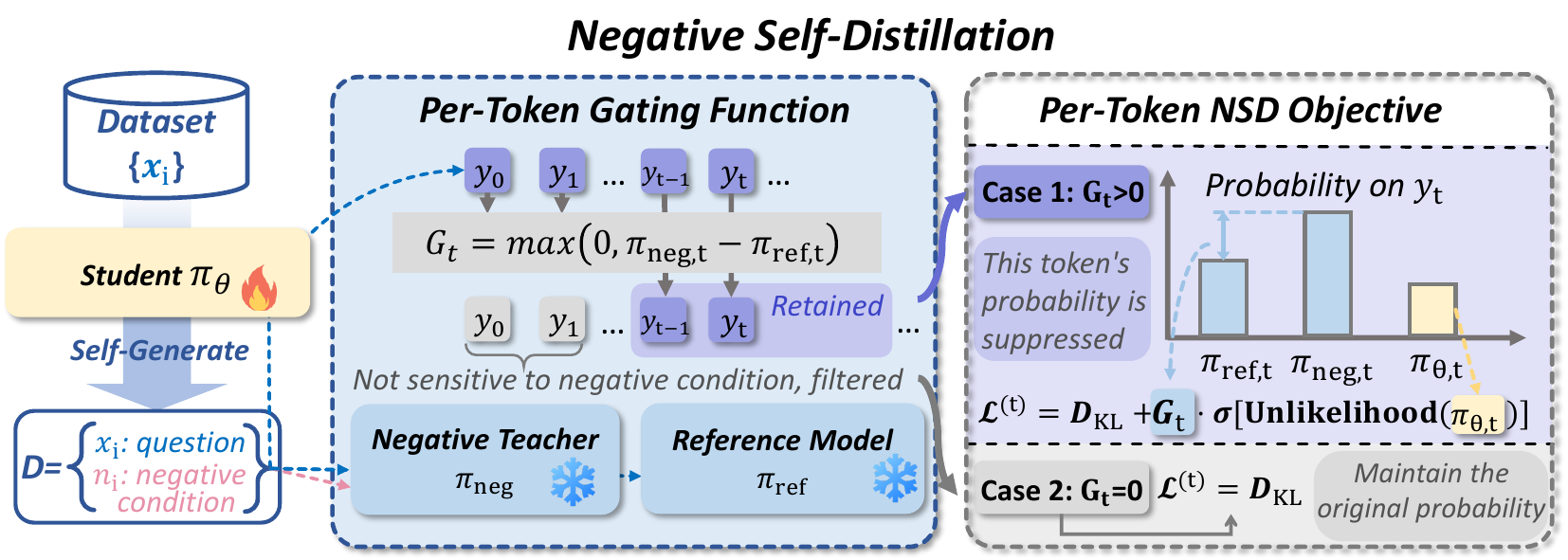} 
    \caption{Overview of Negative Self-Distillation. The student model generates negative conditions from the unlabeled training data (Left). We then compare the token distributions between benign and negative contexts, isolating the tokens whose probabilities are abnormally boosted by the negative condition (Mid). Finally, the model is penalized to suppress the probabilities of these isolated tokens, while the filtered benign tokens are regularized only by KL divergence (Right).}
    \label{fig:method}
\end{figure}

We consider a label-free training dataset denoted as $\mathcal{D}_{\text{raw}} = \{(x_i)\}_{i=1}^N$, where $x_i$ represents the problem statement.
Our method consists of two core components: self negative conditioning and NSD training. We first prompt the student model $\pi_{\theta}$ to generate a negative condition prompt $n_i$ for each problem. By conditioning the model on this prompt $n_i$, we construct a negative teacher $\pi_{\text{neg}}$. We then penalize the student's alignment with the teacher under a simple gating mechanism to avoid applying penalty to reasoning-irrelevant tokens. The overview of NSD is shown in Figure~\ref{fig:method}.

\subsection{Negative Condition Prompt Generation}
\label{Attack Prompt Generation}

The objective of this module is to allocate a negative instruction $n_i$ to each training sample designed to induce flawed reasoning patterns, thereby augmenting the original $\mathcal{D}_\text{raw}$ into a full negative-conditioned dataset $\mathcal{D} = \{(x_i,n_i)\}_{i=1}^N$. 
The negative condition generation strategy should follow the \textit{self-generation} or \textit{easy-to-get} principle, without utilizing any gold answer.
By default, we adopt an online generation strategy: For a given training problem $x$, we first sample an initial solution $y_{\text{init}}$ from the student model $\pi_\theta$. Conditioned on both the problem and this initial response, we then prompt the student model to generate an adaptive negative condition $n$ based on its existing reasoning trace (the complete prompt is provided in Appendix~\ref{app:prompts}.):
\begin{equation}
    y_{\text{init}} \sim \pi_\theta(\cdot \mid x), \quad n \sim \pi_\theta(\cdot \mid x, y_{\text{init}})
\end{equation}
Our framework can naturally accommodate alternative negative condition generation strategies (discussed in Section~\ref{attack-discuss}). 
We default to generating negative conditions on the fly during training as it provides the most stable and effective supervision signal.

\subsection{Background and Challenges in Unlikelihood Training}
\label{sec:background_challenges}

Our motivation of the NSD training objective is to move the student's logits distribution away from the negative teacher model's flawed reasoning behaviors through dense token-level supervision. 
A natural approach to achieve this is standard unlikelihood training (\cite{unlikelihood}), which minimizes the following objective to suppress the probability of undesirable tokens: $$\mathcal{L}_{\text{unlikelihood}} = -\log(1 - \pi_\theta(y_t \mid x_i, y_{<t}))$$ 


However, directly optimizing the objective to distance the student model from the negative teacher's distribution presents two critical challenges and research questions (RQs): 

(1) Indiscriminately treating every highly probable token under the negatively conditioned teacher as a flaw and applying the unlikelihood training is problematic, as ordinary grammatical tokens can appear in both normal and flawed reasoning; unlearning them could easily lead to the degradation of fundamental reasoning capabilities. \textbf{RQ1}: How to identify the tokens that represent genuine reasoning flaws?

(2) This unbounded unlikelihood objective is catastrophic for highly confident, trivial tokens (\eg, punctuation or spaces) --- it triggers loss explosions and overly strong gradient that destabilize training and destroy the model's inherent logic. 
Specifically, as $\pi_\theta \to 1$, the $\mathcal{L}_{\text{unlikelihood}}$ approaches $\infty$ and the gradient approaches the maximum (as detailed in Appendix~\ref{app:gradient}). As a considerable number of tokens have a relatively high probability, this unbounded penalty triggers gradient explosions, also forcing the student to unlearn fixed fundamental linguistic priors (\eg, how to use punctuations) and rapidly update the model parameters in an unstable direction.
\textbf{RQ2}: How to formulate a penalty to avoid gradient and loss explosions for training stability?

\subsection{The NSD Training Objective}
\label{sec:nsd_objective}

\paragraph{Token-level adaptive gating.}
To address RQ1, we propose the gating mechanism to filter out grammatical tokens. During the training phase, the student model generates reasoning trajectories $y = (y_1, \dots, y_T) \sim \pi_\theta$. To construct the gating signals, we instantiate two frozen teacher models based on the same initial student model:
\begin{itemize}[leftmargin=1em]
    \item \textbf{Reference model ($\pi_{\text{ref}}$):} Conditioned only on the original problem $x_i$, predicting the nominal probability $\pi_{\text{ref}}(y_t \mid x_i, y_{<t})$.
    \item \textbf{Negative teacher ($\pi_{\text{neg}}$):} Conditioned on both the problem and the generated negative prompt $n_i$, predicting the negatively biased probability $\pi_{\text{neg}}(y_t \mid x_i, n_i, y_{<t})$. Note that $\pi_{\text{neg}}$ shares the same model weights as $\pi_{\text{ref}}$, differing only by the negative context. 
\end{itemize}

We introduce a simple gating function that compares the probabilities of both models to filter out ordinary linguistic tokens and identify the tokens sensitive to the negative injection. 
For a given student-generated token $y_t\sim \pi_\theta$, the gate is defined as the adjusted positive divergence between the probability of negative and reference model on this token:
\begin{equation}
    G_t = \max\Big(0, \pi_{\text{neg}}(y_t \mid x_i, n_i, y_{<t}) - \pi_{\text{ref}}(y_t \mid x_i, y_{<t})\Big)
\end{equation}
The gate $G_t \in [0, 1]$ acts as an automatic noise filter. If $\pi_{\text{ref}} \geq \pi_{\text{neg}}$, the token is not activated by a negative condition and naturally exempt from penalization, preserving the model's original generative distribution. Conversely, if $\pi_{\text{neg}} > \pi_{\text{ref}}$, it indicates that the negative prompt has boosted the token's likelihood, marking it as a critical target for suppression. 
Crucially, the penalty weight scales proportionally to this positive gap: a larger divergence directly translates to a heavier penalization.

\paragraph{Gated unlikelihood penalty.}
To formulate a mathematically sound penalty (\ie, the second challenge) and address RQ2, after filtering structural noise via the dynamic gate $G_t$, we introduce a Sigmoid-bounded unlikelihood penalty: We squash the penalty using a Sigmoid function, yielding $\frac{1}{2 - \pi_\theta(y_t \mid x_i, y_{<t})}$. 
The Gated Unlikelihood (GU) penalty is formulated as:
\begin{align}
    \mathcal{L}_{\text{GU}}^{(t)} &= G_t \cdot \sigma \Big(-\log\big(1 - \pi_\theta(y_t \mid x_i, y_{<t})\big) \Big) \nonumber \\
    &= G_t \cdot \frac{1}{2 - \pi_\theta(y_t \mid x_i, y_{<t})} \label{eq:push_loss}
\end{align}

\begin{figure}[htbp]
    \centering
    \includegraphics[width=1\textwidth]{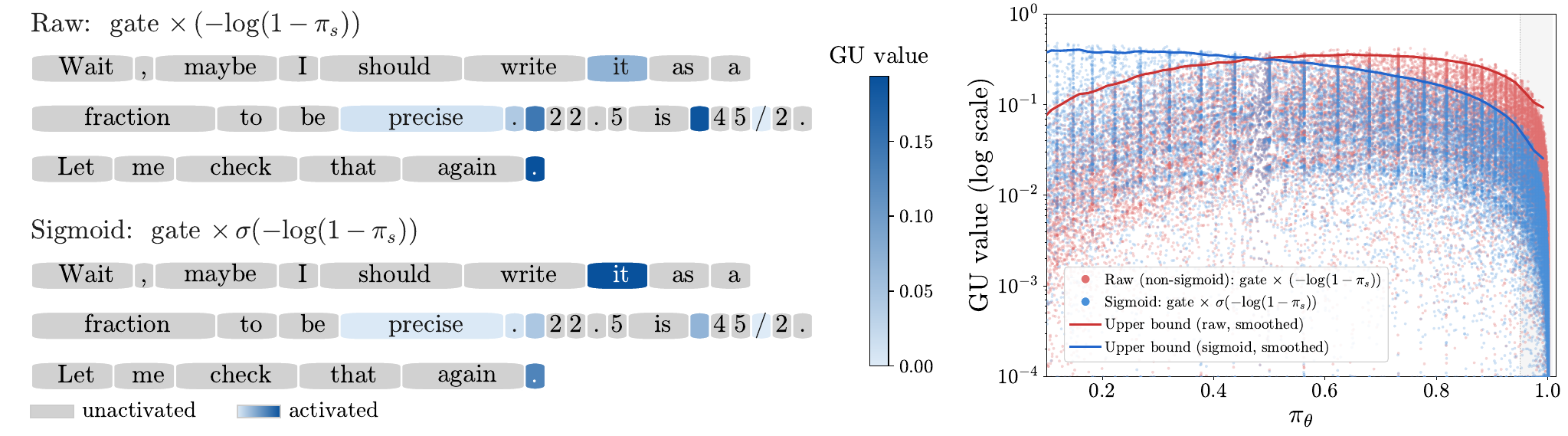} 
    \caption{Left: After applying the sigmoid function, the gated unlikelihood (GU) values are reduced for basic tokens (\eg, punctuations), preventing gradient explosion. Right: The $\mathcal{L}_{\text{GU}}$ value distribution over 4,096 tokens from 100 training samples, showing that the sigmoid objective avoids penalization spikes on high-probability tokens, redistributing the $\mathcal{L}_{\text{GU}}$ weights toward tokens with low-to-mid probabilities in the student model.}
    \vspace{-10pt}
    \label{fig:push-var}
\end{figure}

This bounded formulation actively repels the student from negative flaws while safely preserving essential structural tokens. As shown in Figure~\ref{fig:push-var}, our sigmoid formulation allocates the strongest unlearning signals to low-to-mid confidence tokens, thereby avoiding gradient explosion on high-probability tokens. We further discuss the GU objective in detail in Section~\ref{Push-iv}.

\paragraph{Regularization and overall objective.}
Let $\pi_\theta(y_t \mid x_i, y_{<t})$ denote the current student model being optimized. Our goal is to push the student's distribution away from the identified vulnerabilities without destroying its fundamental linguistic priors. To further regularize the objective, we introduce a point-wise forward KL penalty evaluated on the sampled token $y_t$. Instead of computing the full-vocabulary KL divergence, which is computationally heavy during rollouts, we apply an empirical reference-weighted anchor:
\begin{equation}
    \mathcal{L}_{\text{KL}}^{(t)} = \pi_{\text{ref}}(y_t \mid x_i, y_{<t}) \cdot \log \frac{\pi_{\text{ref}}(y_t \mid x_i, y_{<t})}{\pi_\theta(y_t \mid x_i, y_{<t})}
\end{equation}
This is a single-sample importance-weighted estimator of $D_{\text{KL}}(\pi_{\text{ref}} \| \pi_\theta)$ evaluated on the sampled token $y_t$. 
We finally formulate the NSD loss for a single token $y_t$ as a composite objective:
\begin{equation}
    \mathcal{L}_{\text{NSD}}^{(t)} = \mathcal{L}_{\text{GU}}^{(t)} + \alpha \cdot \mathcal{L}_{\text{KL}}^{(t)} 
\end{equation}
where $\alpha$ is a hyperparameter. The full NSD algorithm is shown in Algorithm~\ref{alg:nsd}.

For every component in the $\mathcal{L}_{\text{NSD}}$, we validate its necessity and effectiveness through ablation studies in Section~\ref{sec:understanding}. The overall objective is calculated by aggregating the token-level losses across the dataset:
\begin{equation}
\label{equal:distill}
\mathcal{J}(\theta) = \mathbb{E}_{(x, n) \sim \mathcal{D}, y \sim \pi_\theta} \left[ \sum_{t=1}^{|y|} \mathcal{L}_{\text{NSD}}^{(t)} \right]
\end{equation}

\begin{algorithm}[htbp]
\caption{Negative Self-Distillation (NSD) Training}
\label{alg:nsd}
\begin{algorithmic}[1]

\Require Unlabeled dataset $\mathcal{D}_{\text{raw}}=\{x_i\}_{i=1}^N$,
Initial model $\pi_{\theta_0}$, KL weight $\alpha$
\Ensure Optimized student model $\pi_\theta$

\State Initialize student $\pi_\theta$, and frozen teachers
$\pi_{\text{ref}},\pi_{\text{neg}}\gets\pi_{\theta_0}$

\For{$x_i\in\mathcal{D}_{\text{raw}}$}
    \State Sample reasoning trajectory
    $y=(y_1,\ldots,y_T)\sim\pi_\theta(\cdot\mid x_i)$
    and the negative prompt
    $n_i\sim\pi_\theta(\cdot\mid x_i,y)$
    \State Initialize loss $\mathcal{J}_i\gets 0$

    \For{$t=1,\ldots,T$}
        \State $p_{\text{ref}}\gets
        \pi_{\text{ref}}(y_t\mid x_i,y_{<t})$
        \State $p_{\text{neg}}\gets
        \pi_{\text{neg}}(y_t\mid x_i,n_i,y_{<t})$
        \State $p_\theta\gets
        \pi_\theta(y_t\mid x_i,y_{<t})$

        \State $G_t\gets\max(0,p_{\text{neg}}-p_{\text{ref}})$
        \State $\mathcal{L}_{\text{NSD}}^{(t)}
        \gets\frac{G_t}{2-p_\theta}
        +\alpha p_{\text{ref}}
        \log\frac{p_{\text{ref}}}{p_\theta}$
        \State $\mathcal{J}_i\gets
        \mathcal{J}_i+\mathcal{L}_{\text{NSD}}^{(t)}$
    \EndFor

    \State Update $\theta$ using gradient
    $\nabla_\theta\mathcal{J}_i$
\EndFor

\State \Return $\pi_\theta$

\end{algorithmic}
\end{algorithm}

\section{Experimental Setup}

\paragraph{Training setup.}
We use the MATH \citep{MATH} dataset as training dataset (for NSD, Intuitor and TTRL training, we discard the gold labels). 
We conduct training on the following models: Qwen3-1.7B, Qwen3-4B, and Qwen3-8B \citep{qwen3technicalreport}.
All models are trained for a total of 2 epochs, which is enough to plateau in all baselines.
We set $ \alpha = 0.01$, top-$k$ = 32, batch size = 32. For NSD, we set the max generation length to 4096.

\paragraph{Evaluation.}
We evaluate the math reasoning ability of all models on the following benchmarks: AIME 2024, AIME 2025, AIME 2026, HMMT 2025~\citep{aime}, MATH-500, AMC 2023 and OlympiadBench~\citep{OlympiadBench}. For OlympiadBench, we exclude the proof problems.
By default, we set hyperparameters according to the recommended setting in Qwen3 report \citep{qwen3technicalreport}: temperature = 0.6; top-$p$ = 0.95; top-$k$ = 20. The output length is set to 32K.

\paragraph{Baselines.} We compare with the following methods representing three different training paradigms:
{\bf OPSD}~\citep{opsd}: A standard distillation framework that minimizes the full-vocabulary KL divergence between the student and a teacher conditioned on the gold solution.
{\bf Intuitor}~\citep{rlif}: A representative RLIF (Reinforcement Learning from Internal Feedback) implementation, which is a variant of GRPO and utilizes average confidence (self-certainty) as the intrinsic reward.
{\bf TTRL}~\citep{ttrl}: A variant of GRPO that utilizes the majority-voting consensus as pseudo-gold labels. While vanilla TTRL typically optimizes directly on the test set, we apply it to the training dataset to ensure a fair comparison with other baseline methods.

A conceptual comparison of NSD with existing related methods, along with their implementation details and prompt templates, is provided in Appendix~\ref{com-other-method} and~\ref{app:exp_details}.

\section{Evaluation Results}

In this section, we first present the main experimental results across multiple mathematical reasoning benchmarks (\S\ref{main-result}). Then we empirically demonstrate NSD achieves better training efficiency and promotes reflection abilities compared to other baselines (\S\ref{nsd-reflection}, \S\ref{nsd-opsd}).
Finally, we show that employing simpler negative conditioning strategies in NSD can also yield comparable effectiveness (\S\ref{attack-discuss}).

\subsection{Main Results}
\label{main-result}
\begin{table*}[htbp]
\centering
\caption{Main evaluation results on mathematical reasoning benchmarks. We report the $\text{Avg@8}$ (\%) performance under non-thinking mode (we report the performance under thinking mode in Appendix~\ref{app:thinking-mode}).
$\Delta$ Avg is the average absolute improvement over the same-size baseline across all 7 benchmarks. We report the best checkpoint on the validation set within 2 training epochs. \textbf{Bold} marks the best result in each model-size group; \underline{underline} marks the second best. $\dagger$ denotes methods that require ground-truth labels. The last two columns report the 95\% CI and one-sided $p$-value (which measures the probability of observing an improvement at least as large as the really observed one if the improvement were due to chance) for $\Delta$ Avg@8, respectively. OlympiadBench is evaluated on the 675 open-ended math problems (excluding proof problems) using the official judger with symbolic comparison.}
\label{tab:main_result}
\resizebox{\textwidth}{!}{
\begin{tabular}{l ccccccc c cc}
\toprule
\textbf{Method} & \makecell{\textbf{AIME} \\ \textbf{2024}} & \makecell{\textbf{AIME} \\ \textbf{2025}} & \makecell{\textbf{AIME} \\ \textbf{2026}} & \makecell{\textbf{HMMT} \\ \textbf{2025 Feb}} & \makecell{\textbf{AMC} \\ \textbf{2023}}  & \makecell{\textbf{Olympiad-} \\ \textbf{Bench}} & \makecell{\textbf{MATH-} \\ \textbf{500}}  & $\mathbf{\Delta}$ \textbf{Avg} & \makecell{\textbf{95\% CI}} & $p$ \\
\midrule

\multicolumn{11}{l}{\textit{1.7B Models}} \\
Qwen3-1.7B & 9.6 & 10.0 & 9.6 & 7.1 & 44.1 & 37.1 & 62.5 & — & — & — \\
OPSD$^\dagger$ & \textbf{15.0} & \underline{14.2} & 8.8 & 5.8 & 44.1 & \underline{37.2} & 62.5 & \underline{+1.1} & $[-0.3, +2.4]$ & $0.06$ \\
Intuitor & 13.8 & 8.3 & 8.3 & 6.7 & 43.4 & 35.4 & 60.6 & $-$0.5 & $[-1.8, +0.8]$ & $0.29$ \\
TTRL & 11.3 & 11.3 & \underline{9.6} & \textbf{8.3} & 41.6 & 36.8 & \textbf{63.4} & +0.3 & $[-1.0, +1.5]$ & $0.29$ \\
\rowcolor{blue!8} NSD & \underline{14.2} & \textbf{17.9} & \textbf{10.0} & \underline{7.1} & \textbf{45.9} & \textbf{38.7} & \underline{62.6} & \textbf{+2.3} & $[+0.7, +4.0]$ & $0.001$ \\
\midrule

\multicolumn{11}{l}{\textit{4B Models}} \\
Qwen3-4B & 23.8 & 20.4 & 17.9 & 10.8 & 68.8 & 47.8 & 71.2 & — & — & — \\
OPSD$^\dagger$ & 25.4 & 22.5 & 15.8 & \underline{15.8} & 68.8 & 47.6 & \underline{71.7} & +1.0 & $[-0.3, +2.4]$ & $0.10$ \\
Intuitor & 24.6 & \underline{25.8} & \underline{18.3} & 13.8 & \underline{70.0} & \underline{47.7} & 69.8 & \underline{+1.3} & $[-0.5, +3.1]$ & $0.05$ \\
TTRL & \underline{25.8} & 19.6 & \underline{18.3} & 11.7 & 68.1 & 47.1 & 71.7 & +0.2 & $[-1.2, +1.7]$ & $0.33$ \\
\rowcolor{blue!8} \textbf{NSD} & \textbf{35.8} & \textbf{31.3} & \textbf{29.2} & \textbf{16.3} & \textbf{76.3} & \textbf{51.0} & \textbf{73.1} & \textbf{+7.5} & $[+5.4, +9.5]$ & $<10^{-4}$ \\
\midrule

\multicolumn{11}{l}{\textit{8B Models}} \\
Qwen3-8B & 28.8 & 19.2 & 18.3 & 11.7 & 67.2 & 48.9 & 73.1 & — & — & — \\
OPSD$^\dagger$ & 30.0 & \underline{21.3} & 17.1 & 12.1 & 66.9 & 48.2 & \underline{73.5} & +0.3 & $[-1.3, +1.9]$ & $0.33$ \\
Intuitor & \underline{34.6} & 20.4 & \underline{18.3} & \underline{13.8} & \underline{70.9} & \underline{49.5} & 72.7 & \underline{+1.9} & $[+0.2, +3.4]$ & $0.02$ \\
TTRL & 29.2 & 18.3 & 17.1 & 10.8 & 69.1 & 49.1 & 73.0 & $-$0.1 & $[-1.4, +1.3]$ & $0.57$ \\
\rowcolor{blue!8} \textbf{NSD} & \textbf{39.6} & \textbf{26.3} & \textbf{25.0} & \textbf{17.9} & \textbf{75.6} & \textbf{50.6} & \textbf{74.1} & \textbf{+6.0} & $[+4.0, +7.9]$ & $<10^{-4}$ \\
\bottomrule
\end{tabular}
}
\end{table*}

The main results are shown in Table~\ref{tab:main_result}. We highlight the following key observations:

\paragraph{NSD achieves the overall best performance on the models with different sizes.} 
As shown in Table~\ref{tab:main_result}, NSD consistently achieves the highest average improvements across all model scales, yielding $\Delta$ Avg gains of \textbf{+2.3\%}, \textbf{+7.5\%}, and \textbf{+6.0\%} on the three models respectively. While baselines like OPSD$^\dagger$ and RL excel narrowly on AIME 2024, our 4B and 8B models achieve broader generalization across diverse math tasks, maintaining peak AIME accuracies of 35.8\% and 39.6\%. Notably, unlike other baselines where small improvements possibly partly stem from randomness, NSD guarantees stable performance gains, supported by a significantly low $p$-value.

\paragraph{NSD is more promising on larger model sizes due to self-generated negative conditions.}
An observation from Table~\ref{tab:main_result} is that NSD exhibits stronger performance gains on larger models compared to the smaller 1.7B variant. 
This scaling behavior is tied to our online negative condition generation mechanism: NSD uses on the model itself to generate solution-specific negative conditions. 
By optimizing against these higher-quality conditions, larger models receive a stronger contrastive training signal, which translates into substantial improvements on challenging reasoning tasks.

\paragraph{Why does NSD outperform other baselines?}

Compared to OPSD, 
label-free training of NSD without the privileged information prevents bias (\eg, reinforcing reasoning shortcuts due to the gold solution) and reflection collapse caused by overconfidence~\citep{opsdfail}.
We provide additional analysis in Section~\ref{nsd-reflection} that further confirms NSD better promotes reflection behaviors than other methods.
Case studies in Appendix~\ref{app:case} also illustrate how NSD-trained models abandon the wrong reasoning trajectory and switch to the right one. 
Compared to Intuitor and TTRL which use model confidence or majority voting to generate training signals,
NSD removes the reliance on the model's self-judgement ability, which leads to possible incorrect training signals. 
For example, weaker models hardly gain improvement from Intuitor (-0.5\% on Qwen3-1.7B) because their high confidence does not necessarily equate to high accuracy. Furthermore, those confidence-based bootstrapping methods also degrade the reflection ability, as shown in Section~\ref{nsd-reflection}.


\subsection{NSD Inspires Reflection}
\label{nsd-reflection}

\textbf{NSD prevents over-confidence and preserves exploratory reflection.} We evaluate model reflection capabilities by measuring the average frequency of reflection tokens (\eg, ``\textit{Wait}'') across AIME and HMMT benchmarks (Table~\ref{tab:confidence_comparison}). 
The detailed definition of reflection tokens is shown in Appendix~\ref{app:task-style}. We observe that OPSD and Intuitor severely suppress reflective behavior (dropping to 2.18 and 0.75 per response, respectively), as training on ground-truth or unverified positive rollouts encourages overly direct, non-verifying reasoning trajectories. 

\vspace{-5pt}
\begin{table}[htbp]
  \centering
  \caption{The average reflection token frequency per response on Qwen3-4B.}
  \label{tab:confidence_comparison}
  \begin{tabular}{lcccc}
      \toprule
      \textbf{Method} & \textbf{AIME 2024} & \textbf{AIME 2025} & \textbf{HMMT 2025} & \textbf{Average} \\
      \midrule
      Baseline & 6.8 & 2.2 & 1.7 & 3.6 \\
      OPSD  & 2.6 & 2.1 & 1.8 & 2.2 \\
      Intuitor & 0.6 & 1.0 & 0.7 & 0.8 \\
      NSD & \textbf{6.9} & \textbf{7.5} & \textbf{8.1} & \textbf{7.5} \\
      \bottomrule
  \end{tabular}
\end{table}
\vspace{-5pt}

Conversely, NSD substantially enhances reflection frequency (yielding up to 7.5 per response). By penalizing flawed reasoning paths, NSD avoids over-confidence and enables the model to autonomously re-evaluate potential errors during complex inference, which is also demonstrated by our case study in Appendix~\ref{app:case}.

\subsection{Negative Condition Variants Study}
\label{attack-discuss}

In our main experiments, we default to an online self negative condition generation strategy (denoted as online strategy briefly). While intuitively well-motivated, this approach incurs computational overhead from online rollouts. To explore more efficient alternatives, we investigate the impact of simpler conditioning strategies, selecting the variants based on effective LLM negative conditioning paradigms identified in \citep{attack_way}. 
In this section, we discuss the following offline generation strategies while our primary evaluations in the previous sections are conducted using the online paradigm:

Strategy 1: Solution-aware negative conditioning. Besides inputting the question, we let the student model rollout first, then prompt it to generate a negative condition based on the question and rollout. 

Strategy 2: Question-only negative conditioning. Input the training sample question to the frozen initial student and prompt it to generate a possible negative condition based on it.

Strategy 3: Noise conditioning. Simply add irrelevant Wikipedia articles as noise (denoted as wiki-irr strategy; ``irr'' stands for irrelevant).

\begin{figure}[htbp]
    \centering
    \includegraphics[width=1\textwidth]{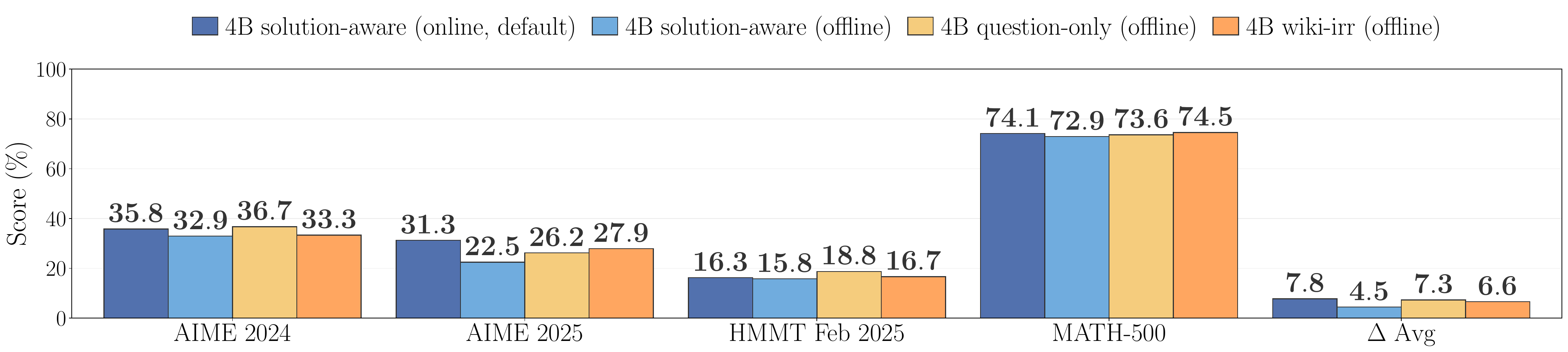} 
    \caption{Evaluation results of NSD across different conditioning strategies. $\Delta$ denotes the average absolute improvement over the base model across these 4 datasets. The dashed line represents the reference $\Delta$ achieved by the default online strategy.}
    \label{fig:atk_results}
\end{figure}

We evaluate the three NSD conditioning strategies on the Qwen3-4B model. 
The experimental result of different conditioning strategies is shown in Figure~\ref{fig:atk_results}. 
Notably, the question-only negative conditioning strategy achieves a 7.3\% average improvement, comparable to 7.8\% using our default online solution-aware approach.
Furthermore, even the most lightweight offline strategy (wiki-irr) also performs competitively with our default approach, highlighting NSD's broad scalability to diverse and efficient negative conditions. 
In contrast, the offline solution-aware strategy exhibits relatively lower performance, primarily driven by its reliance on outdated offline-generated solutions during conditioning.

\begin{wrapfigure}{r}{0.45\textwidth}
  \centering
  \vspace{-4em}
  \includegraphics[width=\linewidth]{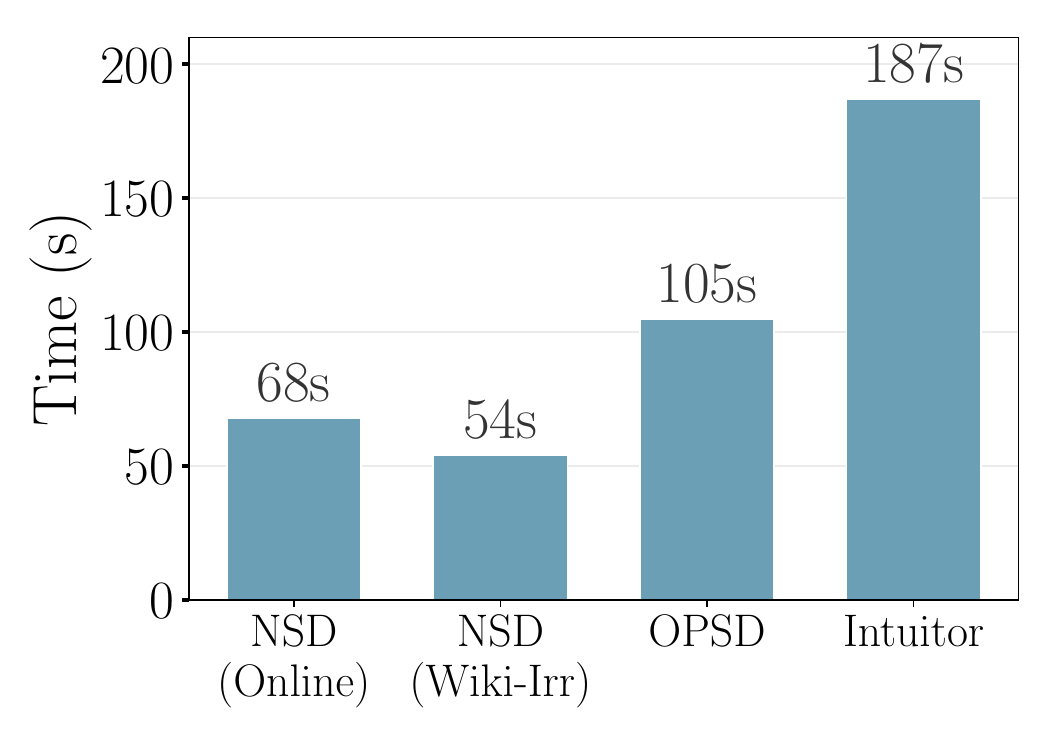} 
  \vspace{-20pt}
  \caption{Average wall-clock time per training step with 6 or 8 A100 GPUs. For NSD and OPSD, the student model occupies 4 GPUs and the teacher occupies 2 GPUs. For Intuitor, the generation stage is executed across all 8 GPUs. Notably, the wiki-irr strategy effectively reduces the latency compared to using the default online rollout in NSD.}
  \vspace{-2em}
  \label{fig:time}
\end{wrapfigure}

\subsection{Efficiency of NSD}
\label{nsd-opsd}

\textbf{NSD exhibits superior training efficiency compared to OPSD and RLIF.}
We focus our detailed latency analysis on the computational overhead of the rollout phase, which is the dominant source of training time discrepancy across different algorithms.

In the rollout stage, the student model samples $\text{batch size} \times n$ rollouts, where $n$ denotes the number of samples per prompt. While GRPO-based baselines (Intuitor and TTRL) demand $n=8$, both NSD and OPSD require only $n=1$. 
Subsequently, OPSD and NSD perform additional forward passes on the generated sequences: OPSD prefills each concatenated prompt-response pair to extract top-$k$ log-probabilities, where $k=32$ in NSD and $128$ in OPSD; online NSD generates an online negative condition based on the student's solution before running two forward passes to compute $\pi_{\text{ref}}$ and $\pi_{\text{neg}}$. 

The time consumption is shown in Figure~\ref{fig:time}. Overall, NSD achieves superior training efficiency through three primary factors:
(1) \textbf{Minimal Rollout Overhead:} Unlike multi-sample GRPO-style baselines, NSD requires only a single rollout per sample, reducing rollout time by $\sim$60\%. Furthermore, static negative condition generation strategies (\eg, wiki-irr) save online rollout time entirely, lowering the overall latency from 68s to 54s.
(2) \textbf{Parallelized Forward Prefilling:} Although computing $\pi_{\text{ref}}$ and $\pi_{\text{neg}}$ involves two distinct prompts, both share the same model weights and can be prefilled concurrently in parallel.
(3) \textbf{Scalar-Only Loss Computation:} NSD requires only three scalar token probabilities, avoiding full-vocabulary logit projections. The result shows that NSD trains faster overall than OPSD, proving that our parallelized prefilling costs substantially less than OPSD's Top-$k$ logit alignment.

\subsection{More Evaluations and Analyses} 
We conduct several supplementary evaluations provided in Appendix~\ref{app:further-exp}. First, we evaluate our models under the Pass@8 metric in Appendix~\ref{app:pass@8}. Second, we report the performance under thinking mode in Appendix~\ref{app:thinking-mode}. NSD continues to outperform all baselines under these settings. Third, in Appendix~\ref{app:sec:pg}, we investigate an alternative objective formulation that treats the negative of the loss as an advantage signal for policy-gradient optimization, demonstrating that the NSD framework is scalable to policy-gradient-style training paradigms.
\section{Understanding NSD Training Objective}
\label{sec:understanding}

\subsection{Adaptive Gating Analysis}
\label{gating}

The adaptive gating function is designed to filter out trivial tokens while retaining essential ones. RLCSD~\citep{rlcsd} rigorously conceptualizes this by categorizing tokens into style and task tokens, treating the former as noise. A detailed definition is shown in Appendix~\ref{app:task-style}. To evaluate how effectively the NSD gating function and existing weighting methods filter out style tokens, we sample a subset of 100 training queries and analyze the logit distributions across the initial 4,096 tokens and calculate the style-task ratio ($\mathcal{T}$ denotes the set of task tokens and $\mathcal{S}$ denotes the set of style tokens):

\begin{equation}
R = {\left[\dfrac{1}{|\mathcal{S}|}\displaystyle\sum_{t \in \mathcal{S}}w_t\right]}/{\left[\dfrac{1}{|\mathcal{T}|}\displaystyle\sum_{t \in \mathcal{T}} w_t\right]}
\end{equation}

We compare NSD adaptive gating with initial ratio, entropy-based OPSD weighting~\citep{8020rule}, and vanilla OPSD loss~\citep{opsd}:

NSD: $w_t = \max\bigl(0,\ \pi_\text{neg}(y_t \mid x, a, y_{<t}) - \pi_\text{ref}(y_t \mid x, y_{<t})\bigr)$;
Entropy-OPSD: $w_t =  -\sum_{v} \pi_\theta(v\mid x, y_{<t})\log \pi_\theta(v
  \mid x, y_{<t})$;
OPSD: $w_t = \sum_{v} \pi_\theta(v)\log\frac{\pi_\theta(v)}{\pi_\text{gold}(v)}$.

A lower $R$ inherently signifies a better approach~\citep{rlcsd}; it implies the method prioritizes task tokens, suppressing gradient generation on meaningless tokens --- previous work~\citep{rlcsd,opsd} shows that in OPSD, the training signal might be dominated by style tokens, causing the student to imitate styles rather than learning reasoning ability.
As shown in Table~\ref{tab:style_task_ratio},
the NSD gating mechanism alone filters style tokens more effectively than both entropy-based and OPSD-loss-based weighting methods.

\begin{table*}[htbp]
    \centering
    \caption{Comparison of style-task ratio across different methods. A lower value indicates a better approach~\citep{rlcsd}.}
    \label{tab:style_task_ratio}
        \begin{tabular}{lccccc}
            \toprule
            \textbf{Method} & \textbf{\makecell{NSD gate \\ (wiki)}} & \textbf{\makecell{NSD gate \\ (solution-aware)}} & \textbf{\makecell{NSD gate \\ (question-only)}} & \textbf{\makecell{Entropy-\\OPSD}} & \textbf{\makecell{OPSD}} \\
            \midrule
            \textbf{style-task ratio} &  2.6$\times$ & 3.4$\times$ & 3.5$\times$ & 3.9$\times$ & 5.4$\times$ \\
            \bottomrule
        \end{tabular}
\end{table*}

\subsection{KL Ablation}
\label{KL-a}

\begin{wrapfigure}{r}{0.5\textwidth}
  \centering
  \vspace{-2em}
  \includegraphics[width=\linewidth]{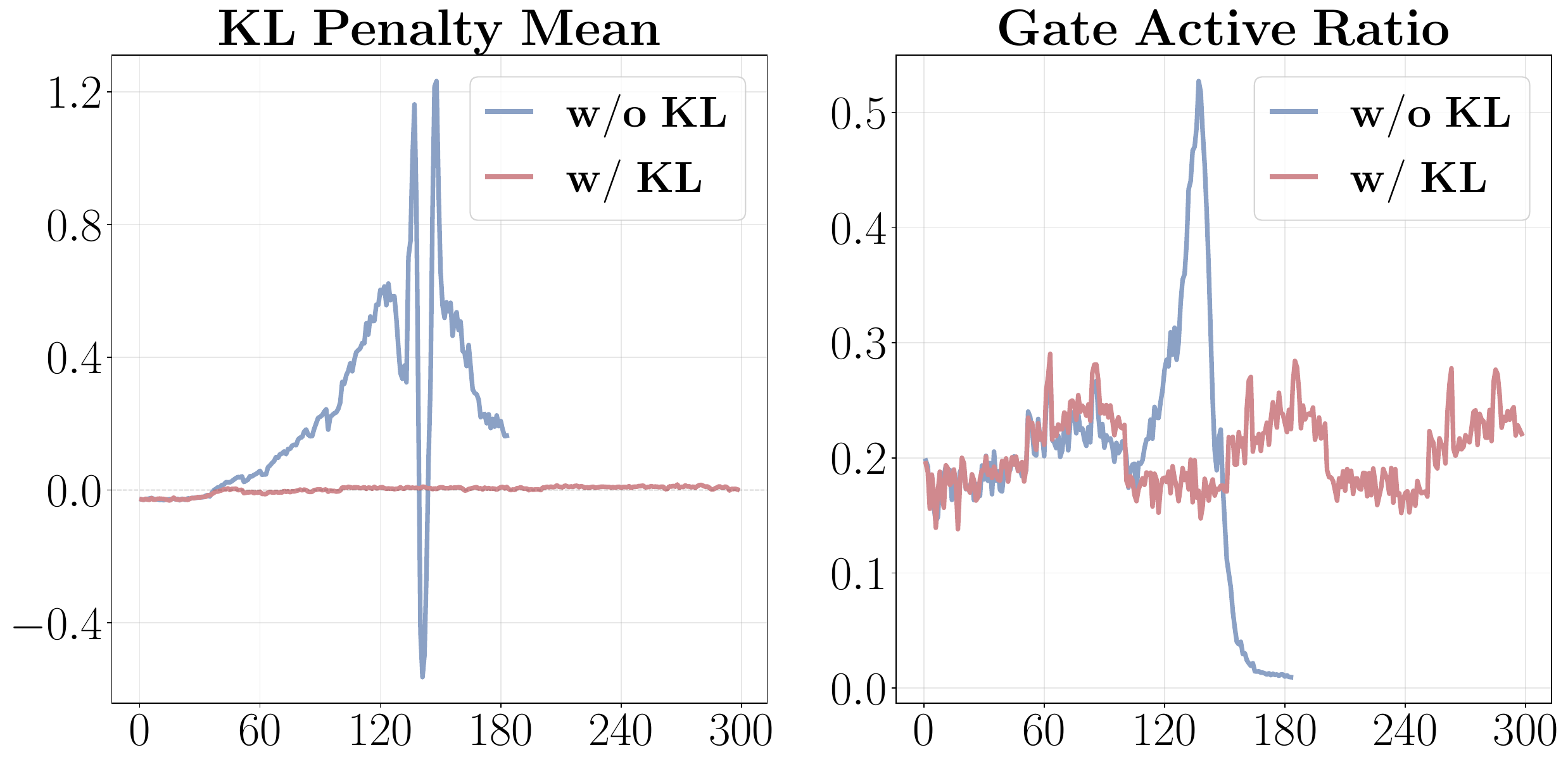} 
  \caption{Training log of NSD w/ and w/o KL constraint on Qwen3-4B. Left: Forward KL between the reference model and student model per training step; Right: The average activated gating ($\overline{G_t}$) per training step.}
  \label{fig:KL}
  \vspace{-1em}
\end{wrapfigure}

We investigate the necessity of the KL divergence constraint within the NSD framework. Figure~\ref{fig:KL} illustrates this via an ablation study comparing the standard NSD against a variant without the KL anchor (NSD-noKL).

Removing the KL constraint leads to a mid-training collapse. As shown in Figure~\ref{fig:KL} (left), NSD-noKL drastically shifts the student's distribution, inducing a cycle of learning and forgetting, evidenced by sharp oscillations in the curve. 
Furthermore, Figure~\ref{fig:KL} (right) demonstrates that the gate activation ratio in NSD-noKL initially increases but drops precipitously around the 120th step, coinciding precisely with the KL collapse. 
We also observe that the mean $\mathcal{L}_{\text{GU}}$ value decreases significantly, indicating that the gating mechanism activates spuriously and loses its effectiveness—a direct result of the model drifting excessively from the reference without KL regularization.

\subsection{Discussion on Gated Unlikelihood}
\label{Push-iv}
A natural inherent consequence of the gating formulation is its sensitivity to minor probability fluctuations in highly predictable tokens (where $\pi_\text{neg} \approx \pi_\text{ref} \to 1$, usually trivial tokens like punctuations). Occasionally, inherent variance may cause the negative teacher model to assign a marginally higher probability than the reference model, bypassing the filter (\eg, probabilities for space tokens often fluctuate slightly around 99\%). 
Nevertheless, this artifact is controlled: the minuscule divergence yields a near-zero gate value $G_t$, ensuring that the overall gating on these high-confidence tokens remains negligible.

To further avoid distancing from these tokens, we bound the pure unlikelihood penalty via Sigmoid function (Equation \ref{eq:push_loss}), so that the model enjoys implicit gradient attenuation. The Sigmoid penalty serves as a structural failsafe against gating imperfections. As shown in Figure~\ref{fig:push-var}, $\mathcal{L}_{\text{GU}}$ value on high-probability tokens is significantly lower than the value of pure unlikelihood objectives.
Gradient analysis is mathematically discussed in Appendix~\ref{app:gradient}.

\section{Related Work}
\label{sec:related}

\textbf{On policy distillation.} The original OPD~\citep{opd,lu2025onpolicydistillation,opd2026survey} relies on external reward models. 
Self distillation~\citep{opsd,sdpo,sdft} removes external teachers by using ground-truth solutions as hints, but suffers from solution bias and overconfidence~\citep{opsdfail,PI-biased,denser-not-better}; Recent studies have increasingly optimized the distillation method across various dimensions, mainly including weak supervision~\citep{sd-zero,opsdwosupervision}, credit assignment~\citep{rlcsd,AgentOPSD,tip,disagreementlearnable}, agentic scenarios~\citep{seed,agenticrlsd}, and other better learning objectives~\citep{oprd,OPD^2,TR-OPD,antisd,Rebellious}.

\textbf{Label-free reinforcement learning.} Existing label-free training methods primarily rely on substituting rewards with self-generated ones (usually based on confidence or entropy) within RLVR frameworks~\citep{rlif,confidenceneed,self-reward,maximizingconfidence,huang2026rzero,huang2026g} or OPD frameworks~\citep{CANON,opsdwosupervision}, as well as generating gold labels by the model itself~\citep{consistent_paths,ttrl}.

\textbf{Training with negative signals.} Unlikelihood objective~\citep{unlikelihood,dontsay} has been proposed to train earlier small language models. Several studies incorporate both positive and negative trajectories into distillation or RLVR frameworks~\citep{xu-etal-2026-harnessing,howdonegative,rlcd}. Notably, NSR~\citep{effectiveNR} explores RLVR training driven exclusively by negative signals, demonstrating that it can preserve high-confidence priors while mitigating overfitting. Furthermore, in the context of self-distillation, recent works introduce negative signals to alleviate student overconfidence~\citep{antisd,Rebellious}.
\section{Conclusion}
\label{sec:conclusion}

In this work, we propose Negative Self-Distillation (NSD), a label-free training framework comprising negative conditioning and gated unlikelihood training. 
Empirical results demonstrate that NSD consistently outperforms existing baselines across seven mathematical reasoning benchmarks under various model sizes. 
Comprehensive analyses and ablation studies show that our adaptive gating mechanism effectively isolates genuinely flawed tokens, while the sigmoid unlikelihood objective ensures smoother reasoning gradients. 
Furthermore, NSD accommodates diverse negative conditioning strategies, establishing it as a highly scalable framework. 
Its training efficiency is enhanced by bypassing full-vocabulary computations and leveraging parallelized forward passes for the negative teacher and reference model. 
Importantly, NSD inherently preserves and stimulates the model's capacity for self-reflection, highlighting its potential as a promising post-training method for enhancing the reasoning capabilities of LLMs. 
\section*{Limitations}

NSD relies on the student model's inherent capacity to generate negative conditions. Consequently, this approach may be less effective for extremely small or weak models that struggle to produce meaningful negative contrasts for optimization. 
However, given the rapid capability scaling of modern foundational models, this capacity bottleneck is expected to diminish naturally in future architectures or in stronger models.

Under the online strategy, NSD requires negative-condition generation and two forward passes through the two same frozen models. 
Nevertheless, we explored alternative conditioning strategies, including efficient generation-free methods like the wiki-irr strategy, which can mitigate the rollout costs while maintaining competitive performance.
Moreover, executing the two forward passes in parallel at each training step effectively minimizes overall wall-clock latency.

\section*{Acknowledgments}
This research is partially funded by the NVIDIA Academic Grant and Amazon Research Award.
We thank Xinyu Wang and Yu Gu for their valuable feedback and suggestions, particularly for the experimental design.

\bibliography{iclr2027_conference}
\bibliographystyle{iclr2027_conference}

\newpage
\appendix
\section{Analysis of Candidate Gated Unlikelihood Gradients in NSD and OPSD Objectives}
\label{app:gradient}

The NSD loss function is defined as $\mathcal{L} = \alpha \cdot D_{\text{KL}}(\pi_{\text{ref}} \parallel \pi_{\theta}) + \mathcal{L}_{\text{GU}}$. Specifically, we focus on isolating and analyzing the $\mathcal{L}_{\text{GU}}$ term, which penalizes the student model on tokens vulnerable to negative conditions. Let $\pi_c =   \pi_{\theta}(c)$ denote the student's predicted probability for the target sampled token, and $G = \max(0, \pi_{\text{neg}} - \pi_{\text{ref}})$ serve as the adaptive gate. To demonstrate the necessity of our $\mathcal{L}_{\text{GU}}$ item with Sigmoid design, we compare two distinct candidates for this penalty:
\begin{align}
    \text{Standard Unlikelihood:} \quad & \text{GU}_{\text{std}} = G \cdot [-\log(1 - \pi_c)] \label{eq:push_std} \\
    \text{Ours:} \quad & \text{GU}_{\text{sig}} = G \cdot \sigma(-\log(1 - \pi_c)) = G \cdot \frac{1}{2 - \pi_c} \label{eq:push_sig}
\end{align}
The parameter update magnitude is driven by the gradient of the loss with respect to the pre-softmax logit, $\frac{\partial \text{GU}}{\partial z_c}$. Given the logit-probability Jacobian $\frac{\partial \pi_c}{\partial z_c} = \pi_c(1 - \pi_c)$, we evaluate the optimization behavior of both formulations below.

\begin{figure}[htbp]
    \centering
    \includegraphics[width=1\textwidth]{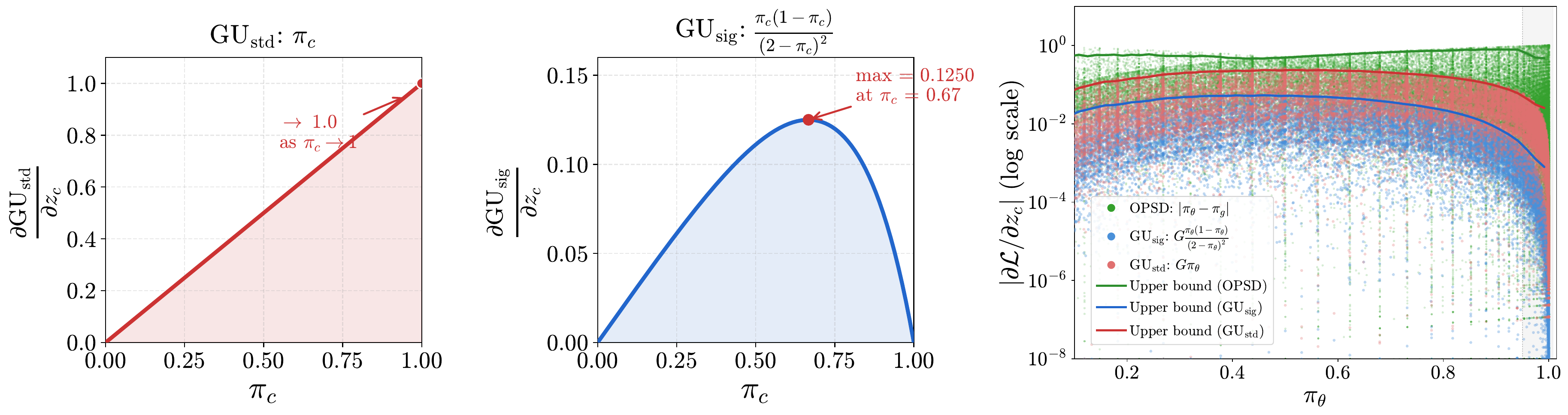} 
    \caption{(Left and Mid) Comparison of the variations of two types of gradients by probability. (Right) The real gradient distribution in 100 training samples. OPSD tends to assign larger gradients to high-probability tokens, making the model more prone to drastic updates. In contrast, compared to the vanilla unlikelihood loss, our GU objective further suppresses the gradients on high-probability tokens.}
    \label{fig:app:compare}
\end{figure}

\subsection{Gradient Hazard in Standard Unlikelihood}
Applying the chain rule to the standard unbounded logarithmic penalty (Equation \ref{eq:push_std}), the gradient with respect to the logit is:
\begin{equation}
    \frac{\partial \text{GU}_{\text{std}}}{\partial z_c} = G \cdot \frac{1}{1 - \pi_c} \cdot \pi_c(1 - \pi_c) = G \cdot \pi_c
\end{equation}

Mathematically, the gradient in standard unlikelihood scales strictly linearly with the student's confidence $\pi_c$ (as shown in the Figure~\ref{fig:app:compare}). This creates an optimization hazard: In causal language modeling, tokens with extreme confidence ($\pi_c > 0.9$) are probably trivial structural tokens---such as fixed collocations, prepositions, and punctuation. 
Under this formulation, whenever the gate $G$ is triggered when $\pi_c \to 1$, the optimizer delivers its almost maximum update magnitude to these hyper-confident function words. 
This aggressively penalizes the model's fundamental linguistic priors, leading to a degradation in generation fluency, especially when the high-probability token ratio is high per rollout.

\subsection{Implicit Gradient Attenuation in Sigmoid-Squashed GU}
To construct a noise-resilient supervision signal, our method utilizes the Sigmoid-squashed penalty (Equation \ref{eq:push_sig}). Deriving the logit gradient for this formulation yields:
\begin{align}
    \frac{\partial \text{GU}_{\text{sig}}}{\partial z_c} &= G \cdot \frac{1}{(2 - \pi_c)^2} \cdot \pi_c(1 - \pi_c) \nonumber \\
    &= G \cdot \frac{\pi_c(1 - \pi_c)}{(2 - \pi_c)^2}
\end{align}

This formulation introduces an elegant, parameter-free implicit gradient attenuation mechanism. The presence of the $(1 - \pi_c)$ term in the numerator fundamentally alters the gradient landscape. As the student model's probability approaches $1$, the gradient magnitude decays toward zero:
\begin{equation}
    \lim_{\pi_c \to 1} \frac{\partial \text{GU}_{\text{sig}}}{\partial z_c} = 0
\end{equation}

Since $\pi_c > 0.9$ predominantly corresponds to uninformative syntactic tokens, the Sigmoid function inherently protects the model's structural fluency by silencing huge gradient on these tokens. 
Instead, as shown in Figure~\ref{fig:app:compare} it naturally concentrates the highest gradient magnitude on mid-confidence tokens ($\pi_c \approx 0.6$), which are more likely to be the ambiguous, reasoning-critical tokens where the student model requires the strongest corrective supervision. 
Consequently, our squashed formulation guarantees that dense supervision remains targeted and stable.

\subsection{OPSD Gradient Analysis}

OPSD objective can be described by: 

\begin{equation}
    \mathcal{L}_{\text{OPSD}}^{(t)} = D_{\text{KL}} \Big(\pi_{\theta}(\cdot \mid x_i, s_i, y_{<t}) \parallel \pi_\theta(\cdot \mid x_i, y_{<t}) \Big)
\end{equation}

$s_i$ denotes the gold solution in $i$-th training sample. The gradient visualization is shown in Figure~\ref{fig:app:compare}. This figure shows that, compared with the NSD objective, whose gradient generally decreases as the token probability increases, the OPSD objective exhibits an increasing trend. This indicates that OPSD encourages the model to learn more from high-probability tokens, which may lead to certain forms of reward hacking, such as overlearning style tokens. In contrast, the candidate objectives in NSD exhibit relatively stable gradient patterns, while the sigmoid-based GU can more effectively suppress gradients on high-probability tokens.

\section{Comparison of NSD with Other Methods}
\label{com-other-method}

\begin{table}[htbp]
\centering
\renewcommand{\arraystretch}{1.2}
\caption{Comparison of NSD with other methods. We conceptually compare them across the following dimensions: \textbf{Sampling} denotes whether the training relies on trajectories generated by the model itself; \textbf{Source of reward signal} indicates the core component driving the training loss function; \textbf{Teacher} specifies whether the approach depends on an external teacher model; \textbf{Gold label} refers to whether ground-truth answers are required; and \textbf{Monitor signal quality} represents whether the method actively filters training signals (\eg, unconsciously or intentionally) rather than indiscriminately optimizing over all tokens.
Note that our NSD is a label-free approach, which is not directly comparable to baselines that rely on additional or external supervision. Consequently, our main experiments mostly focus on comparable methods, with OPSD as a representative label-dependent method for reference.}
\resizebox{\linewidth}{!}{
\begin{tabular}{llllll}
\toprule
\textbf{Method} 
& \textbf{Sampling} 
& \makecell[l]{\textbf{Source of} \\ \textbf{reward signal}} 
& \textbf{Teacher} 
& \textbf{Gold label} 
& \makecell[l]{\textbf{Monitor signal} \\ \textbf{quality}} \\
\midrule

\makecell[l]{SFT/Off-Policy \\ Distillation} 
& \bad\ off-policy 
& external teacher  
& \bad\ external 
& \good\ no     
& \bad\ no \\

RLVR (GRPO)   
& \good\ on-policy  
& gold label 
& \good\ no       
& \bad\ needed 
& \makecell[l]{\good\ noise gradients are \\ \good\ counteracted} \\

\makecell[l]{OPD}      
& \good\ on-policy  
& external reward  
& \bad\ external 
& \good\ no     
& \bad\ no \\

OPSD/SDPO     
& \good\ on-policy  
& gold label  
& \good\ self     
& \bad\ needed     
& \bad\ no \\

SD-Zero       
& \good\ on-policy  
& trained reviser  
& \good\ self     
& \bad\ needed     
& \bad\ no \\

RLIF       
& \good\ on-policy  
& internal metric  
& \good\ no     
& \good\ no     
& \bad\ no \\

TTRL/U-OPSD       
& \good\ on-policy  
& majority-voting  
& \good\ no     
& \good\ no     
& \bad\ no \\

\midrule

\textbf{NSD}  
& \good\ on-policy  
& negative condition  
& \good\ self     
& \good\ no     
& \makecell[l]{\good\ noise is filtered by \\ gating} \\

\bottomrule
\end{tabular}%
}
\end{table}

\section{Experiment Details}\label{app:exp_details}

\subsection{Hyperparameters}
\label{app:hyper}

Table~\ref{tab:hyper_all} lists the training hyperparameters for all methods.
All experiments are conducted on a single node equipped with 8 NVIDIA A100 (80GB) GPUs. Unless otherwise specified, we adopt the default hyperparameters from the respective official implementations, with the following controlled adjustments for fair comparison: For OPSD, we evaluate configurations both with and without LoRA and report the best-performing variant (where LoRA achieves superior results on the 1.7B and 4B models). For Intuitor, we standardize the training batch size to 128, deviating from their scale-dependent defaults (64 for smaller models and 128 for larger models). For TTRL, as majority voting relies on complete final solutions, we extend the maximum generation length to 8192 tokens to prevent output truncation.

\begin{table*}[htbp]
\centering
\caption{Training hyperparameters for all methods.
  ``---'' means not applicable.}
\label{tab:hyper_all}
\resizebox{\textwidth}{!}{
\begin{tabular}{l l l l l}
\toprule
\textbf{Hyperparameter} & \textbf{NSD (Online, Solution-aware)} & \textbf{OPSD} & \textbf{Intuitor} & \textbf{TTRL} \\
\midrule
GPUs                  & 4 for actor + 2 for teacher & 4 for actor + 2 for teacher & 8                 & 8 \\
Train batch size      & 32                & 32                & 128               & 8 \\
PPO mini-batch size   & 32                & 32                & 128               & 1 \\
Max prompt length     & 512               & 512               & 512               & 512 \\
Max response length   & 4096              & 4096              & 3072              & 8192 \\
Actor learning rate   & $1\times10^{-6}$  & $5\times10^{-6}$  & $3\times10^{-6}$  & $5\times10^{-7}$ \\
LR warmup ratio       & 0.1               & 0.1               & 0.1               & 0.03 \\
Rollout per sample $n$           & 1                 & 1                 & 8                 & 8 \\
Top-$k$ logits        & 32                & -1              & ---               & --- \\
KL coefficient        & 0.01              & ---               & 0.005             & 0.00 \\
Total epochs          & 2                 & 2                 & 2                 & 2 \\
\midrule
\multicolumn{5}{l}{\textit{OPSD LoRA target modules: all-linear, with LoRA rank = 64 and alpha = 128.}} \\
\bottomrule
\end{tabular}
}
\end{table*}

\subsection{Templates}
\label{app:template}

We use the Qwen3 instruct chat template throughout.
All training are conducted in \textbf{non-thinking mode}: the chat template is invoked with \texttt{enable\_thinking=False}, which causes the model to emit an empty \texttt{<think>} block and proceed directly to the answer.
This applies uniformly to the student rollout, the teacher log-probability computation, and all downstream evaluations.

The template for a single-turn exchange takes the following form:

\begin{tcolorbox}[
  colback=gray!6, colframe=gray!50, arc=4pt,
  title={\small\texttt{Qwen3 Chat Template (non-thinking, \texttt{enable\_thinking=False})}},
  fonttitle=\bfseries\small,
  left=6pt, right=6pt, top=4pt, bottom=6pt
]
\small
\begin{verbatim}
<|im_start|>user
{user message}
<|im_end|>
<|im_start|>assistant
<think>

</think>

{model response}
\end{verbatim}
\end{tcolorbox}

\noindent
The empty \texttt{<think>\ldots</think>} block is prepended automatically by the template when \texttt{enable\_thinking=False} and \texttt{add\_generation\_prompt=True}.
The model then generates its response after the second blank line.

\subsection{Prompts}
\label{app:prompts}

The student always receives the plain problem prompt below.
During NSD training the teacher receives either the same prompt (reference pass) or a negative prompt (negative pass), depending on the variant.
All prompts are wrapped in the chat template described in Appendix~\ref{app:template}.

\textbf{Student / reference teacher prompt (all methods).}

\begin{tcolorbox}[
  colback=gray!6, colframe=gray!40, arc=4pt,
  title={\small\texttt{Student Prompt}}, fonttitle=\bfseries\small,
  left=6pt, right=6pt, top=4pt, bottom=4pt
]
\small
\texttt{Problem: \{problem\}}\\[4pt]
\texttt{Let's think step by step and output the final answer within \textbackslash boxed\{\}.}
\end{tcolorbox}

\paragraph{NSD negative condition prompt generator (Question-only).}
The following meta-prompt is sent to a helper LLM to produce the per-sample negative condition prompt $n_i$ used in the question-only offline variant.
The generated prompt replaces the system context seen by the teacher model.

\begin{tcolorbox}[
  colback=blue!4, colframe=blue!30, arc=4pt,
  title={\small\texttt{Meta-Prompt: Question-only Negative Condition Generation}}, fonttitle=\bfseries\small,
  left=6pt, right=6pt, top=4pt, bottom=4pt
]
\small
You are an expert Math Educator and AI Prompt Engineer.
Your task is to analyze the following math problem and generate a ``Generalized Attack Prompt'' that will force an LLM to make a highly plausible, human-like cognitive error.

\medskip
\textbf{Anatomy of a Universal Attack Prompt:}
\begin{enumerate}[leftmargin=1.4em, itemsep=1pt, topsep=2pt]
  \item \textbf{Persona:} Must start exactly with \emph{``You are a student who\ldots''}.
        Describe a specific bad habit relevant to \emph{this} problem.
  \item \textbf{Trigger:} Abstract the problem's mathematical class.
        \emph{Never} use specific numbers or variables from the current problem.
  \item \textbf{Flawed Execution:} Instruct a naive heuristic or impulsive shortcut that would give a wrong answer.
  \item \textbf{Fatal Omission:} Explicitly forbid the critical verification step.
\end{enumerate}

\medskip
Now, perform this task for the following problem:\\[2pt]
\textbf{Problem:} \{problem\}

\medskip
Output \emph{only} the ``Generalized Attack Prompt''.
Start your response with \emph{``You are a student who\ldots''}.
Keep it concise (2--3 sentences).
\end{tcolorbox}

\paragraph{NSD negative condition prompt generator (Solution-aware).}
When the model's own rollout is available, the meta-prompt is augmented with the student's solution to produce a more targeted negative condition.

\begin{tcolorbox}[
  colback=orange!5, colframe=orange!35, arc=4pt,
  title={\small\texttt{Meta-Prompt: Solution-aware Negative Condition Generation}}, fonttitle=\bfseries\small,
  left=6pt, right=6pt, top=4pt, bottom=4pt
]
\small
You are an expert Math Educator and AI Prompt Engineer.
Your task is to analyze the following math problem and a student's existing solution, then generate a ``Targeted Attack Prompt'' that exploits the exact reasoning steps the student used to cause a highly plausible cognitive error.

\medskip
The student's solution reveals \emph{how} they solved this problem---use that to craft an attack targeting their specific reasoning steps.

\medskip
\textbf{Anatomy of a Targeted Attack Prompt:}
\begin{enumerate}[leftmargin=1.4em, itemsep=1pt, topsep=2pt]
  \item \textbf{Persona:} Must start exactly with \emph{``You are a student who\ldots''}.
        Describe a specific bad habit that would corrupt the \emph{exact} step where this student's reasoning is most fragile.
  \item \textbf{Trigger:} Reference the \emph{type} of reasoning the student used
        (not specific numbers or variables from this problem).
  \item \textbf{Flawed Execution:} Instruct a shortcut that mirrors the student's approach but introduces a subtle error.
  \item \textbf{Fatal Omission:} Forbid the specific verification the student performed correctly.
\end{enumerate}

\medskip
\textbf{Problem:} \{problem\}\\[4pt]
\textbf{Student's Existing Solution:}\\
\{solution\}

\medskip
Output \emph{only} the ``Targeted Attack Prompt''.
Start your response with \emph{``You are a student who\ldots''}.
Keep it concise (2--3 sentences).
\end{tcolorbox}   

\paragraph{NSD teacher prompt (wiki-irr variant).}
In the wiki-irr variant no meta-prompt generator is used.
Instead, each training sample is paired with a randomly sampled Wikipedia passage that is concatenated as spurious ``context''.
The teacher sees the following prompt while the student still receives the plain student prompt above.

\begin{tcolorbox}[
  colback=green!4, colframe=green!35, arc=4pt,
  title={\small\texttt{Teacher Prompt: Wiki Irrelevant Negative Condition}}, fonttitle=\bfseries\small,
  left=6pt, right=6pt, top=4pt, bottom=4pt
]
\small
\texttt{Problem: \{problem\}}\\[4pt]
\texttt{Below is some context you may find useful to answering the question above:}\\[4pt]
\texttt{\{wikipedia\_passage\}}\\[4pt]
\texttt{Let's think step by step and output the final answer within \textbackslash boxed\{\}.}
\end{tcolorbox}

\paragraph{NSD teacher prompt.}
For the question-only and solution-aware offline variants, the teacher receives the following prompt, where \texttt{\{negative\_condition\}} is the output of the meta-prompt generator above.

\begin{tcolorbox}[
  colback=red!4, colframe=red!30, arc=4pt,
  title={\small\texttt{Teacher Prompt: Question-only / Solution-aware Negative Condition}}, fonttitle=\bfseries\small,
  left=6pt, right=6pt, top=4pt, bottom=4pt
]
\small
\texttt{Problem: \{problem\}}\\[4pt]
\texttt{\{negative\_condition\}}\\[4pt]
\texttt{Now solve the problem following this instruction:}\\[4pt]
\texttt{Let's think step by step and output the final answer within \textbackslash boxed\{\}.}
\end{tcolorbox}

\section{Additional Experimental Results}
\label{app:further-exp}

\subsection{Pass@8 Performance}
\label{app:pass@8}

We report the performance of pass@8 in Table~\ref{tab:pass8_result}.

\begin{table*}[htbp]
\centering
\caption{Main evaluation results reported as $\text{pass@8}$ (\%): at least one of 8 sampled solutions is correct. Same evaluation setting as Table~\ref{tab:main_result}. $\Delta$ Avg is the average absolute improvement over the same-size baseline across all 7 benchmarks. \textbf{Bold} marks the best result in each model-size group; \underline{underline} marks the second best. $\dagger$ denotes methods that require ground-truth labels.}
\label{tab:pass8_result}
\resizebox{\textwidth}{!}{
\begin{tabular}{l ccccccc c}
\toprule
\textbf{Method} & \makecell{\textbf{AIME} \\ \textbf{2024}} & \makecell{\textbf{AIME} \\ \textbf{2025}} & \makecell{\textbf{AIME} \\ \textbf{2026}} & \makecell{\textbf{HMMT} \\ \textbf{2025 Feb}} & \makecell{\textbf{AMC} \\ \textbf{2023}}  & \makecell{\textbf{Olympiad-} \\ \textbf{Bench}} & \makecell{\textbf{MATH-} \\ \textbf{500}}  & $\mathbf{\Delta}$ \textbf{Avg} \\
\midrule

\multicolumn{9}{l}{\textit{1.7B Models}} \\
Qwen3-1.7B & 16.7 & 23.3 & 13.3 & 16.7 & 70.0 & 57.8 & 77.6 & — \\
OPSD$^\dagger$ & \textbf{40.0} & 23.3 & 13.3 & 16.7 & 72.5 & \underline{59.6} & 77.6 & \underline{+3.9} \\
Intuitor & 30.0 & 23.3 & 16.7 & 13.3 & 75.0 & 57.0 & 76.2 & +2.3 \\
TTRL & 30.0 & \underline{26.7} & \underline{23.3} & 13.3 & \underline{77.5} & 58.7 & 77.0 & +4.4 \\
\rowcolor{blue!8} NSD & \underline{33.3} & \textbf{36.7} & \textbf{23.3} & \textbf{16.7} & \textbf{77.5} & \textbf{60.9} & \textbf{77.6} & \textbf{+7.2} \\
\midrule

\multicolumn{9}{l}{\textit{4B Models}} \\
Qwen3-4B & 50.0 & 40.0 & 40.0 & 20.0 & 95.0 & 67.0 & 81.6 & — \\
OPSD$^\dagger$ & 40.0 & 46.7 & 36.7 & \underline{30.0} & 92.5 & 66.4 & 81.6 & +0.0 \\
Intuitor & 50.0 & \underline{53.3} & 36.7 & 26.7 & \underline{95.0} & 66.1 & 80.0 & +2.0 \\
TTRL & \underline{63.3} & 40.0 & \underline{46.7} & 23.3 & 90.0 & 64.3 & 81.6 & +2.2 \\
\rowcolor{blue!8} \textbf{NSD} & \textbf{60.0} & \textbf{63.3} & \textbf{53.3} & \textbf{30.0} & \textbf{95.0} & \textbf{68.3} & \textbf{82.0} & \textbf{+8.3} \\
\midrule

\multicolumn{9}{l}{\textit{8B Models}} \\
Qwen3-8B & 56.7 & 30.0 & 43.3 & 23.3 & 92.5 & 66.8 & 82.0 & — \\
OPSD$^\dagger$ & 46.7 & \underline{43.3} & 40.0 & 23.3 & 87.5 & 67.6 & 81.8 & $-$0.6 \\
Intuitor & \underline{60.0} & 40.0 & 36.7 & \underline{26.7} & \underline{95.0} & \underline{69.0} & \underline{81.8} & +2.1 \\
TTRL & 56.7 & 33.3 & 36.7 & 20.0 & 92.5 & 66.8 & 82.0 & $-$1.0 \\
\rowcolor{blue!8} \textbf{NSD} & \textbf{70.0} & \textbf{46.7} & \textbf{60.0} & \textbf{40.0} & \textbf{95.0} & \textbf{69.0} & \textbf{81.8} & \textbf{+9.7} \\
\bottomrule
\end{tabular}
}
\end{table*}

\subsection{Performance on Thinking Mode}
\label{app:thinking-mode}

We evaluate the performance of all methods under the thinking mode on the Qwen3-4B model, reporting the results of the best-performing checkpoints evaluated under the thinking mode. As shown in Table~\ref{tab:Think}, NSD also outperforms all other baselines overall. 
Notably, both OPSD and NSD achieve more improvements on challenging datasets such as AIME and Olympiad Bench, aligning with the observations from the non-thinking setting. 
On datasets with limited headroom for improvement (\eg, AMC), all methods perform comparably to the base model. Furthermore, we observe that the Intuitor-trained model tends to over-think, causing many responses to exceed the maximum generation length limit (even after extending it to 38k tokens), which leads to a severe degradation in accuracy. 
This phenomenon is also discussed in the previous works~\citep{rlif,PRISM}.

\begin{table*}[htbp]
\centering
\caption{Thinking mode evaluation results on 4B models reported as avg@8 (\%). Same evaluation setting as Table 1. }
\begin{tabular}{lcccccccc}
\toprule
Model & \makecell{AIME \\ 2024} & \makecell{AIME \\ 2025} & \makecell{AIME \\ 2026} & \makecell{HMMT \\ 2025} & \makecell{AMC \\ 2023} & \makecell{MATH- \\ 500} & \makecell{Olympiad \\ Bench} & \makecell{$\Delta$ \\ Avg} \\
\midrule
Qwen3-4B & 75.8 & 69.1 & \underline{67.5} & 46.0 & \underline{97.2} & 79.8 & 45.9 & - \\
OPSD     & \underline{76.2} & \underline{69.8} & 67.2 & \underline{46.2} & 96.6 & \textbf{80.0} & \underline{46.7} & \underline{+0.2} \\
Intuitor & 52.9 & 45.8 & 51.2 & 39.6 & 90.9  & 78.1 & 43.9 & -11.3 \\
TTRL     & 72.5 & 64.3 & 65.1 & 46.0 & 96.6 & 79.2 & 44.4 & -1.9 \\
\rowcolor{blue!8} NSD & \textbf{77.3} & \textbf{73.3} & \textbf{67.7} & \textbf{48.4} & \textbf{97.8} & \underline{79.9} & \textbf{57.9} & \textbf{+3.0} \\
\bottomrule
\end{tabular}
\label{tab:Think}
\end{table*}

\subsection{Alternative Objective: Policy Gradient Optimization}
\label{app:sec:pg}

While the NSD loss $\mathcal{L}_{\text{NSD}}^{(t)}$ can be directly backpropagated as a supervised objective, we find it also fits a sampled-token advantage policy-gradient framework, following the spirit of~\cite{opsd} and \cite{lu2025onpolicydistillation}.
For each token $y_t$ in a student rollout $y \sim \pi_\theta(\cdot \mid x_i)$, we define a token-level advantage as the NSD loss:
\begin{equation}
    A_t = -\mathcal{L}_{\text{NSD}}^{(t)}
\end{equation}
Intuitively, a token with high NSD loss receives a strongly negative advantage, signaling the policy to reduce its probability. Conversely, tokens with low NSD loss receive near-zero or positive advantage, leaving their probabilities unchanged.
We treat $A_t$ as a constant with respect to $\theta$ and optimize the student via the standard policy gradient surrogate objective:
\begin{equation}
\label{equal:pg}
\mathcal{J}_{\text{PG}}(\theta) =  \mathbb{E}_{(x, a) \sim \mathcal{D}, y \sim \pi_\theta} \left[ \sum_t A_t \log \pi_\theta(y_t \mid x_i, y_{<t}) \right]
\end{equation}

We evaluate the NSD based on the alternative objective under the same setting as our main experiment. The result is shown in Table~\ref{tab:pg_results}. Compared to models optimized with the $\mathcal{J}$ objective (Eq.~\ref{equal:distill}), NSD trained under $\mathcal{J}_{\text{PG}}$ (Eq.~\ref{equal:pg}) achieves superior performance on the 1.7B model (+5.0\% on average). However, on the 4B and 8B models, the $\mathcal{J}$-objective NSD yields better overall results. 
Notably, the wiki-irr strategy consistently performs best under the policy gradient setting, while the online solution-aware strategy emerges as the second best. This discrepancy arises because the gradients are truncated by the advantage function, decreasing the capture of richer gradient signals. 
In contrast, wiki-irr utilizes noise to introduce more generalized interference (causing an overall degradation of the model's reasoning capabilities in long contexts), which ultimately makes it a more effective strategy in this regime.

\begin{table*}[htbp]
\centering
\caption{Evaluation results of NSD with different negative conditioning strategies on mathematical reasoning benchmarks. The models are trained based on the NSD policy gradient objective in Eq.~\ref{equal:pg}.}
\label{tab:pg_results}
\resizebox{\textwidth}{!}{
\begin{tabular}{l cccc c}
\toprule  
\textbf{Method / Variant} & \textbf{AIME 2024} & \textbf{AIME 2025} & \textbf{HMMT Feb 2025} & \textbf{MATH-500} & $\mathbf{\Delta}$ \textbf{Avg} \\
\midrule

\multicolumn{6}{l}{\textit{1.7B Models}} \\
NSD (Solution-aware, Offline) & 15.8 & 14.2 & 5.4 & 63.2 & +2.4 \\
NSD (Solution-aware, Online) & 15.8 & 12.5 & 7.5 & 63.3 & +2.5 \\
\textbf{NSD (Wiki-irr, Offline)} & \textbf{20.0} & \underline{15.0} & \textbf{9.6} & \textbf{64.4} & \textbf{+5.0} \\
\midrule

\multicolumn{6}{l}{\textit{4B Models}} \\
NSD (Question-only, Offline) & 30.4 & 24.6 & 16.2 & 72.7 & +4.4 \\
NSD (Solution-aware, Offline) & \textbf{33.8} & 22.5 & 14.6 & 72.1 & +4.2 \\
NSD (Solution-aware, Online) & 31.3 & 25.0 & 16.3 & \textbf{74.0} & \underline{+5.1} \\
NSD (Wiki-irr, Offline) & \underline{33.8} & \textbf{25.4} & \textbf{17.1} & \underline{73.5} & \textbf{+5.9} \\
\midrule

\multicolumn{6}{l}{\textit{8B Models}} \\
NSD (Solution-aware, Offline) & 30.8 & 23.3 & 12.5 & 73.6 & +1.9 \\
NSD (Solution-aware, Online) & \textbf{35.0} & 21.7 & 13.8 & 73.6 & +2.8 \\
\textbf{NSD (Wiki-irr, Offline)} & \underline{34.2} & \textbf{28.8} & \textbf{19.2} & \underline{73.9} & \textbf{+5.8} \\
\bottomrule
\end{tabular}
}
\end{table*}

\subsection{Case Study}
\label{app:case}

A case study is shown in Table~\ref{tab:case_study}. These results indicate that NSD-trained models more readily explore novel and correct solutions that are entirely absent from the outputs of both the base and OPSD-trained models. 
Furthermore, by prompting an external LLM (Sonnet) to analyze the reasoning traces of each response, we observe that the NSD-trained model engages in several reflection steps, successfully circumventing erroneous trajectories that commonly trap the base model.

\begin{table}[htbp]
\centering
\caption{Case study on AIME 2025 II \#12 comparing Qwen3-4B baseline, OPSD, and NSD. The table shows a summary from Sonnet. Numbers denote correct samples out of 8 independent draws.
  \textcolor{green!60!black}{\ding{51}}~correct; \textcolor{red}{\ding{55}}~incorrect.}
\label{tab:case_study}
\renewcommand{\arraystretch}{1.4}
\resizebox{\textwidth}{!}{%
\begin{tabular}{p{3.8cm} p{4.5cm} p{4.5cm} p{4.5cm}}
\toprule
\textbf{Problem} & \textbf{Baseline} & \textbf{OPSD} & \textbf{NSD (Ours)} \\
\midrule

\multicolumn{4}{p{18.3cm}}{%
  \textit{AIME 2025 II \#12 --- Geometry.}
  Let $A_1A_2\cdots A_{11}$ be a non-convex simple 11-gon satisfying:
  (1)~$[A_iA_1A_{i+1}]=1$ for $2\le i\le10$;
  (2)~$\cos(\angle A_iA_1A_{i+1})=\tfrac{12}{13}$ for $2\le i\le10$;
  (3)~perimeter $= 20$.
  Express $A_1A_2+A_1A_{11}=\tfrac{m\sqrt{n}-p}{q}$ ($n$ squarefree, no prime divides all of $m,p,q$);
  find $m+n+p+q$.
} \\[4pt]

\textbf{Pass@8}
  & 0/8 \quad \textcolor{red}{\ding{55}}
  & 0/8 \quad \textcolor{red}{\ding{55}}
  & \textbf{4/8} \quad \textcolor{green!60!black}{\ding{51}} \\[6pt]

\textbf{Key reasoning}
&
\small
From $\cos\theta=\tfrac{12}{13}$ derives $\sin\theta=\tfrac{5}{13}$, hence
$|A_1A_i|\cdot|A_1A_{i+1}|=\tfrac{26}{5}$.
The product constraint gives an alternating sequence
$a_2=x,\; a_3=\tfrac{26}{5x},\; a_4=x,\;\ldots$
Attempts to use the perimeter but \textit{conflates the sum of radii from $A_1$
with the polygon perimeter}, obtaining $5x+\tfrac{26}{x}=20$ which has no clean
closed form.
After extensive numerical trials, \textbf{guesses the symmetric solution}
$x=\tfrac{13}{\sqrt{5}}$ (i.e.\ $a_2=a_{10}$), giving
$A_1A_2+A_1A_{11}=\tfrac{13}{\sqrt5}+2\sqrt5=\tfrac{23\sqrt5}{5}$,
so $m=23,n=5,p=0,q=5$.
\newline\textbf{Final: $\mathbf{33}$} \quad \textcolor{red}{\ding{55}}
&
\small
Same product relation and alternating sequence.
Applies Law of Cosines: since $x_ix_{i+1}=\tfrac{26}{5}$,
each inner-polygon side satisfies $d^2=x_i^2+x_{i+1}^2-\tfrac{48}{5}$,
so all 9 inner sides are equal.
Correctly writes the perimeter equation $a+9d+\tfrac{26}{5a}=20$
and sets $S=a+\tfrac{26}{5a}$.
\textit{Instead of solving for $S$, minimises $S$ via AM--GM}:
$\min\!\bigl(a+\tfrac{26}{5a}\bigr)=2\sqrt{\tfrac{26}{5}}=\tfrac{2\sqrt{130}}{5}$,
and incorrectly \textbf{treats this minimum as the answer}, concluding
$A_1A_2+A_1A_{11}=\tfrac{2\sqrt{130}}{5}$,
so $m=2,n=130,p=0,q=5$.
\newline\textbf{Final: $\mathbf{137}$} \quad \textcolor{red}{\ding{55}}
&
\small
Same product relation and alternating sequence.
Applies Law of Cosines; all 9 inner sides equal $d$.
Writes the perimeter equation $a+9d+\tfrac{26}{5a}=20$.
Then \textbf{attempts a symmetric-guess approach}: tests
$x=\sqrt{26/5}$, $x=2$, $x=13/5$, $x=13/\sqrt{5}$ in turn,
each time verifying numerically that the two expressions for $d^2$
do \emph{not} agree.
Sets $S=a+\tfrac{26}{5a}$, so the perimeter equation gives $d=\tfrac{20-S}{9}$.
Rewrites $d^2$ via Law of Cosines:
$d^2 = a^2+\tfrac{676}{25a^2}-\tfrac{48}{5}
      = \bigl(a+\tfrac{26}{5a}\bigr)^2-\tfrac{52}{5}-\tfrac{48}{5}
      = S^2-20$.
Substituting $d=\tfrac{20-S}{9}$ yields
$\bigl(\tfrac{20-S}{9}\bigr)^2=S^2-20$,
which expands to $4S^2+2S-101=0$.
Quadratic formula: $S=\tfrac{-2\pm\sqrt{1620}}{8}=\tfrac{9\sqrt5-1}{4}$ (positive root),
so $m=9,n=5,p=1,q=4$.
\newline\textbf{Final: $\mathbf{19}$} \quad \textcolor{green!60!black}{\ding{51}}
\\[4pt]

\textbf{Reflection}
&
\small
\textit{No self-correction.}
After the perimeter approach yields no clean form,
the model \textbf{commits to a guess} ($x=\tfrac{13}{\sqrt5}$)
without checking whether it satisfies the original constraints,
and submits the result directly.
&
\small
\textit{No self-correction.}
The model sets up the correct equation structure
but \textbf{replaces the constraint with its relaxation}:
once the AM--GM bound is computed it is treated as the solution,
with no attempt to verify that the minimum is actually attained.
&
\small
After the guessing strategy fails on multiple candidates
($x=\!\sqrt{26/5}$, $2$, $\tfrac{13}{5}$, $\tfrac{13}{\sqrt5}$),
the model \textbf{explicitly abandons the approach}
(``\textit{Hmm.\ Maybe my approach is not working.\ Alternative idea:}'')
and \textbf{reframes the problem} around the aggregate variable
$S=a+\tfrac{26}{5a}$, turning an intractable system into a
single quadratic $4S^2+2S-101=0$.
\\

\bottomrule
\end{tabular}}
\end{table}

\section{Definition of Task, Style, and Reflection Tokens}
\label{app:task-style}

Considering the similar task setting with RLCSD~\citep{rlcsd}, we follow theie definition of both task and style tokens:

\begin{enumerate}
    \item[(1)] empty or whitespace-only $\rightarrow$ \textbf{style};
    
    \item[(2)] matches any of the math regexes (a digit \texttt{\textbackslash d}; an arithmetic operator in $+ - = * / < > \times \div \leq \geq \neq$; a LaTeX command \texttt{\textbackslash[A-Za-z]+}; a double backslash; or one of \texttt{\$} \verb|^| \verb|_|) $\rightarrow$ \textbf{task};
    
    \item[(3)] the normalized form is in the math wordlist \{\texttt{mod, prime, factor, gcd, lcm, log, ln, sin, cos, tan, exp, integral, sqrt, boxed, frac, sum, prod, pi, alpha, beta, gamma, theta, delta, lambda, mu, sigma, infty, leq, geq, neq, cdot, times, div}\} $\rightarrow$ \textbf{task};
    
    \item[(4)] pure punctuation or a literal newline token (\texttt{\textbackslash n}, \texttt{\textbackslash\textbackslash n}) $\rightarrow$ \textbf{style};
    
    \item[(5)] the normalized form is in the discourse wordlist (connectives \texttt{therefore, so, thus, hence, then, because, since}; hedges \texttt{wait, maybe, perhaps, seems, okay, ok, well, now, first, next, finally, actually, alternatively, however}; scaffolding \texttt{step, answer, let, lets}; closed-class function words \texttt{is, are, us, we, the, a, an, of, to, for, in, on, by, at, as, and, or, but, if, yes, no, this, that, these, those, it, its, be, been, being, have, has, had, do, does, did, will, would, should, could, can, may}) $\rightarrow$ \textbf{style};
    
    \item[(6)] otherwise $\rightarrow$ \textbf{neutral}.
\end{enumerate}

The following tokens are considered as reflection tokens:

\texttt{wait, actually, hmm,
let me reconsider, let me rethink,
i made an error, i made a mistake,
that's wrong, that is wrong,
incorrect, reconsider, rethink, re-examine,
let me check, let me verify,
double check, double-check,
going back, revisit, on second thought}

\end{document}